\documentclass[10pt]{article}

\usepackage[margin=1in]{geometry}
\usepackage{amsmath,amssymb}
\usepackage{booktabs}
\usepackage{graphicx}
\usepackage{microtype}
\usepackage[numbers,sort&compress]{natbib}
\usepackage{xcolor}
\usepackage[hidelinks]{hyperref}
\usepackage[capitalise,nameinlink,noabbrev]{cleveref}

\makeatletter
\def\input@path{{./}}
\makeatother

\title{TANGO: Treating Tokens as Operators}
\author{
Joshua Nunley\\
Department of Informatics\\
Luddy School of Informatics, Computing, and Engineering\\
Cognitive Science Program\\
Indiana University Bloomington\\
\texttt{joshnunl@iu.edu}
}
\date{}

\begin{document}
\newcommand{\SelectedTestCompleteRuns}{288}
\newcommand{\SelectedTestCompleteCells}{96}
\newcommand{\SelectedTestCompletePanels}{16}
\newcommand{\SelectedTestFineWebRuns}{72}
\newcommand{\SelectedTestLeanRuns}{72}
\newcommand{\SelectedTestDMMathRuns}{72}
\newcommand{\SelectedTestStackRuns}{72}

\maketitle
\begin{abstract}
Transformers separate cross-token mixing in self-attention from token-wise
transformation in feed-forward networks. We ask whether combining these
operations can lower predictive loss under fixed data and parameter budgets.
To do so, we introduce the
Token-Aggregated Nonlinear Gating Operator (TANGO) model. TANGO computes a
nonlinear feature-wise gate at each source token. Attention averages these
gates for each destination. The average modulates a linear projection of the
destination and forms the diagonal core of a source-conditioned linear
operator.

We test this proposal by comparing full-prefix and windowed TANGO with looped
and untied Transformers, the Gated Attention Unit (GAU), and Fast Linear
Attention with a Single Head (FLASH) on web text, Lean formal mathematics,
DeepMind Mathematics, and code. The comparison uses two parameter scales, two
depths, and three seeds. Checkpoints are selected on development data and
evaluated on held-out test data. At matched parameters and training data,
full-prefix TANGO has the lowest mean test negative log-likelihood in all 16
settings. In eight additional combinations of size and dataset, its development
loss never increases as depth rises from 4 to 8 to 16, whereas the looped
Transformer's loss increases in four.

Full-prefix TANGO is computationally expensive because it averages wide gates
over every visible source. To reduce this cost, we evaluate a variant with
three narrower gated-projection sets assigned to the first, middle, and last
applications. Across four FineWeb-Edu settings, this variant achieves 3.26 to
3.45 times the throughput of TANGO and 75\% to 96\% that of the looped Transformer.
Its mean development negative log-likelihood is lower than TANGO's in three
settings and 0.023 higher in the fourth, while remaining lower than both
Transformer baselines in all four.
\end{abstract}

\section{Introduction}
\label{sec:introduction}

The Transformer is the dominant architecture for language
modeling~\citep{vaswani2017attention}.  In the standard Transformer, each block
alternates between two operations.  Self-attention mixes information across
tokens, and a position-wise feed-forward network then transforms each token on
its own.  Many alternatives preserve this division while reducing the
computational cost of token mixing.  Sparse and windowed attention
restrict which pairs of tokens interact~\citep{beltagy2020longformer,zaheer2020bigbird},
while linear attention and state-space models summarize the prefix with a
running state~\citep{katharopoulos2020transformers,choromanski2021performer,gu2022efficiently,gu2024mamba}.
A more closely related approach combines the two operations.  The Gated
Attention Unit (GAU) and its linear-time extension, Fast Linear Attention with
a Single Head (FLASH), replace attention and the feed-forward network with one
gated update~\citep{hua2022transformer}.

Training computation has grown faster than the amount of high-quality public
text available, and current projections indicate that the largest training
runs may use all of this text within the next several
years~\citep{villalobos2024will}.
Repeating data remains useful for a limited number of epochs, after which its
value falls quickly~\citep{muennighoff2023scaling}.  Together, these constraints
motivate a complementary question.  When training data and parameter count are
fixed, what predictive loss does each architecture achieve, and how much
computation does it require?  To study this question, we compare six
architectures under matched data and parameter budgets and evaluate their
computational requirements.

Our proposed update combines token mixing with token-wise transformation by
modifying the feed-forward network.  In the Swish-gated linear unit (SwiGLU)
used by modern Transformers, a linear projection of a token is multiplied by a
nonlinear gate computed from that same token~\citep{shazeer2020glu}.  The
proposed update instead lets the gate depend on every causally visible token.
We call the token being transformed the destination.  We call each causally
visible token that contributes a gate a source, including the destination
itself.  Each source produces a nonlinear gate.  Attention averages the source
gates for each destination, and the average modulates a linear projection of
the destination.  With the sigmoid linear unit (SiLU) as the gate activation,
the contribution of a source $x_j$ to a destination $x_i$ is
\begin{equation}
 f(x_i;x_j)=W_o\!\left[
   \operatorname{SiLU}(W_gx_j)\odot W_vx_i\right].
 \label{eq:intro-source-operator}
\end{equation}
Because the gate from source $j$ multiplies the destination's features one
coordinate at a time, it forms the diagonal core of a linear operator between
two shared dense projections.  Each source therefore specifies a linear
operator, and the attention weights form a weighted combination for each
destination.  In this sense, standard attention adds retrieved vectors to the
destination representation, whereas the proposed update applies a retrieved
linear operator.  This construction is related to fast-weight memories and
hypernetworks, in which one part of a network produces the weights that another
part applies~\citep{schmidhuber1992learning,schlag2021linear,ha2017hypernetworks}.
We call the resulting network a Token-Aggregated Nonlinear Gating Operator
(TANGO) model.

The model reuses the same block parameters at every application, as a looped
Transformer does.  Within this shared-block architecture, we evaluate two
attention patterns.  Full-prefix TANGO attends to every visible source.
Windowed TANGO uses the same scores within a recent window and summarizes older
tokens with running sums.  This windowed computation makes its cost linear in
sequence length and, at our longest context of 8,192 tokens, brings its
arithmetic close to that of a looped Transformer.

To test whether the proposed update improves predictive accuracy under matched
data and parameter budgets, our primary comparison spans four domains.  These
domains are educational web text from FineWeb-Edu, Lean source files and
proofs, DeepMind Mathematics, and code in six programming languages from The
Stack v3.  Among these domains, Lean is closest to the data-limited regime
because its training corpus is small and our training budgets repeat it many
times.  We train six architectures at two sizes of about 70 million (70M) and
160 million (160M) parameters, with four or eight applications of the block and
three paired seeds.  Each run's checkpoint is selected on development data and
then evaluated once on a held-out test set.  Our contributions are the
following.
\begin{itemize}
 \item We introduce TANGO, a token-mixing update in which nonlinear source
 features define the diagonal core of a linear map at each destination.
 We also introduce a windowed variant whose cost is linear in sequence length.
 \item We compare TANGO with five alternatives over 288 training runs.  TANGO
 has the lowest mean test negative log-likelihood (NLL) in all 16
 combinations of dataset, size, and depth at matched parameters and training
 data, while the ranking of the other architectures changes from setting to
 setting.  The largest standard deviation for TANGO is 0.045 NLL, compared
 with 0.053 to 0.319 for the non-TANGO baselines.  In
 additional runs at depth 16, TANGO's development loss never rises as depth
 increases from 4 to 8 to 16, whereas that of the looped Transformer rises in
 half of the settings.
 \item We evaluate TANGO with three narrower sets of gated projections for
 the first, middle, and last applications.  Across four FineWeb-Edu settings,
 its throughput is 3.26 to 3.45 times that of TANGO and 75\% to 96\% that of
 a looped Transformer.  Its mean development NLL is lower than that of TANGO
 in three settings and 0.023 higher in the fourth.  It remains lower than both
 Transformer baselines in every setting.
\end{itemize}

\section{Model}
\label{sec:method}

\begin{figure}[t]
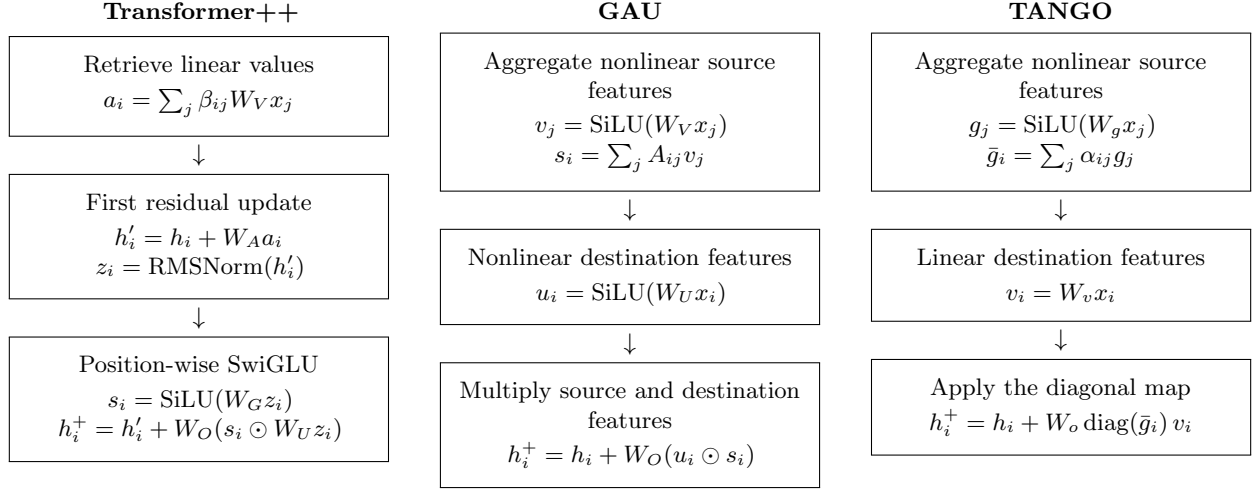

  \centering
  \setlength{\fboxsep}{7pt}
  \begin{minipage}[t]{0.31\textwidth}
    \centering\small
    \textbf{Transformer++}\\[4pt]
    \fbox{\parbox{0.88\linewidth}{\centering
      Retrieve linear values\\[2pt]
      $a_i=\sum_j\beta_{ij}W_Vx_j$}}
    \\[2pt]$\downarrow$\\[2pt]
    \fbox{\parbox{0.88\linewidth}{\centering
      First residual update\\[2pt]
      $h'_i=h_i+W_Aa_i$\\
      $z_i=\operatorname{RMSNorm}(h'_i)$}}
    \\[2pt]$\downarrow$\\[2pt]
    \fbox{\parbox{0.88\linewidth}{\centering
      Position-wise SwiGLU\\[2pt]
      $s_i=\operatorname{SiLU}(W_Gz_i)$\\
      $h_i^+=h'_i+W_O(s_i\odot W_Uz_i)$}}
  \end{minipage}\hfill
  \begin{minipage}[t]{0.31\textwidth}
    \centering\small
    \textbf{GAU}\\[4pt]
    \fbox{\parbox{0.88\linewidth}{\centering
      Aggregate nonlinear source features\\[2pt]
      $v_j=\operatorname{SiLU}(W_Vx_j)$\\
      $s_i=\sum_j A_{ij}v_j$}}
    \\[2pt]$\downarrow$\\[2pt]
    \fbox{\parbox{0.88\linewidth}{\centering
      Nonlinear destination features\\[2pt]
      $u_i=\operatorname{SiLU}(W_Ux_i)$}}
    \\[2pt]$\downarrow$\\[2pt]
    \fbox{\parbox{0.88\linewidth}{\centering
      Multiply source and destination features\\[2pt]
      $h_i^+=h_i+W_O(u_i\odot s_i)$}}
  \end{minipage}\hfill
  \begin{minipage}[t]{0.31\textwidth}
    \centering\small
    \textbf{TANGO}\\[4pt]
    \fbox{\parbox{0.88\linewidth}{\centering
      Aggregate nonlinear source features\\[2pt]
      $g_j=\operatorname{SiLU}(W_gx_j)$\\
      $\bar g_i=\sum_j\alpha_{ij}g_j$}}
    \\[2pt]$\downarrow$\\[2pt]
    \fbox{\parbox{0.88\linewidth}{\centering
      Linear destination features\\[2pt]
      $v_i=W_vx_i$}}
    \\[2pt]$\downarrow$\\[2pt]
    \fbox{\parbox{0.88\linewidth}{\centering
      Apply the diagonal map\\[2pt]
      $h_i^+=h_i+W_o\operatorname{diag}(\bar g_i)\,v_i$}}
  \end{minipage}
  \caption{Update structures, with $x_i$ the normalized input and head indices
  and positional transformations suppressed.  Both GAU and TANGO multiply an
  attention aggregate of nonlinear source features by destination features.
  The destination branch is nonlinear in GAU and linear in TANGO.  Conditional
  on its routing weights and source aggregate, the TANGO update is therefore a
  linear map with a diagonal core.  TANGO also weights sources with
  multihead normalized cosine scores $\alpha_{ij}$ and a zero-valued null option
  rather than GAU's squared rectified-linear scores $A_{ij}$.  The vertical
  arrangement is only for readability.}
  \label{fig:block-comparison}
\end{figure}

\paragraph{Gated update.}
Let $h_i\in\mathbb{R}^d$ be the residual state at position $i$ and
$x_i=\operatorname{RMSNorm}(h_i)$, where RMSNorm denotes root mean square
layer normalization.  Each position produces a nonlinear source gate and
linear destination features,
\begin{equation}
 g_i=\operatorname{SiLU}(W_gx_i)\in\mathbb{R}^{F},\qquad
 v_i=W_vx_i\in\mathbb{R}^{F}.
 \label{eq:cross-gate-value}
\end{equation}
We split each gate into $H$ heads $g_{ia}\in\mathbb{R}^{F/H}$, and we
write $d_h=d/H$ for the query and key dimension of each head.  The query and
key of head $a$ are $q_{ia}=\operatorname{RoPE}_i(W_{q,a}x_i)$ and
$k_{ia}=\operatorname{RoPE}_i(W_{k,a}x_i)$, where RoPE denotes rotary position
embedding~\citep{su2021roformer}, and we normalize each query and key to unit
length.  For head $a$ and a visible source $j\leq i$, the logit is
$\ell_{ija}=\tau_a q_{ia}^{\top}k_{ja}$ with learned temperature
$\tau_a=\exp(\min(\theta_a,\log 20))$.  Let $\mathcal C_i=\{1,\ldots,i\}$ be
the set of causal sources.  Given nonnegative source masses $m_{ija}$, the
aggregated gate and the residual update are
\begin{equation}
 \bar g_{ia}=
 \frac{\sum_{j\in\mathcal C_i}m_{ija}g_{ja}}
 {\sum_{j\in\mathcal C_i}m_{ija}+|\mathcal C_i|e^{b_a}},\qquad
 h_i^+=h_i+W_o\!\left(v_i\odot\operatorname{concat}_a\bar g_{ia}\right),
 \label{eq:tango-update}
\end{equation}
where $\operatorname{concat}_a$ concatenates the aggregated heads.  The learned
scalar $b_a$ assigns a mass of $e^{b_a}$ to the null option for every visible
source.  Because the null option carries a zero gate, it provides a learned
source-independent way to reduce the update magnitude.
\Cref{fig:block-comparison} compares this update with GAU and with our main
Transformer baseline, Transformer++.

\paragraph{Tokens as operators.}
Writing $\bar g_i=\operatorname{concat}_a\bar g_{ia}$, the update in
\cref{eq:tango-update} becomes
\begin{equation}
 h_i^+=h_i+W_o\operatorname{diag}(\bar g_i)\,W_vx_i.
 \label{eq:tokens-as-operators}
\end{equation}
For fixed attention weights and source gates, the gated part of the update is a
linear map of the destination projection.  Its diagonal core is assembled from
the source gates weighted for that destination.  Because the SiLU gate is unbounded and can be
negative, this core can amplify a feature or flip its sign as well as suppress
it.  In fast-weight terminology, the gate at each source is a diagonal fast
weight, and each destination query retrieves a weighted mixture of these fast
weights (\cref{sec:related}).  The same update
can also be read as multihead attention whose values are the width-$F$ gates
$g_j$, followed by elementwise multiplication with the destination features
$v_i$.  This second reading makes the cost of the model explicit, because each
attention step now moves vectors of width $F$ rather than $d$.

\paragraph{Full-prefix and windowed aggregation.}
Full-prefix TANGO sets $m_{ija}=\exp(\ell_{ija})$, which gives a
softmax-like average over the source gates and the null option.  For a
sequence of length $T$, computing the scores costs $\Theta(T^2d)$ and averaging
the $F$-dimensional gates costs $\Theta(T^2F)$.  Because $F$ is much larger than
$d$ in our parameter-matched models, the second term dominates.

Windowed TANGO keeps the same unnormalized exponential scores for the most
recent $W=64$ sources.  For older sources it uses a positive feature map built
from the exponential linear unit (ELU),
\begin{equation}
 m_{ija}=\phi(\sqrt{\tau_a}q_{ia})^\top
          \phi(\sqrt{\tau_a}k_{ja}),\quad
 \phi(u)=\frac{\max(\operatorname{ELU}(u)+1+10^{-4},10^{-4})}{\sqrt{d_h}}.
 \label{eq:window-score}
\end{equation}
Because this mass factorizes over queries and keys, the contributions of older
sources can be computed from running sums of the transformed keys and of their
outer products with the source gates, and recent and older masses share the
denominator in \cref{eq:tango-update}.  This older-source weighting rule is
learned directly rather than constrained to approximate the full-prefix
weights.  Its cost is linear in $T$ when $W$, $d$, $F$, and the
number of applications are fixed.

\paragraph{Repeated application and parameter sharing.}
The primary TANGO models tie the block parameters across $K$ depth steps.
\begin{equation}
 \mathbf h^{(r+1)}=\operatorname{Block}_{\vartheta}
 (\mathbf h^{(r)}),\qquad r=0,\ldots,K-1.
 \label{eq:cross-recurrence}
\end{equation}
TANGO shares the normalization parameters, query and key projections,
temperatures, null biases, and the projections $W_g$, $W_v$, and $W_o$ across
all applications.  A variant of TANGO instead stores three sets of $W_g$,
$W_v$, and $W_o$ while continuing to share the other parameters.  The first
application uses the initial set, the next $K-2$ applications reuse the middle
set, and the last application uses the final set.  We call this model TANGO
with three gated-projection sets.  At the same number of stored parameters,
each of its three sets is narrower than the single set in TANGO.  Its width is
$F=9{,}488$ rather
than $28{,}480$ at 70M, and $F=17{,}400$ rather than $52{,}176$ at 160M.  Since
only one set runs at each application, the active gated width falls by roughly
a factor of three.

\paragraph{Baselines.}
We use Transformer++ to mean a Transformer with pre-RMSNorm, rotary
self-attention, and SwiGLU feed-forward networks, following the modern
components summarized by \citet{gu2024mamba}.  Loop Transformer++ applies one
such block $K$ times, and Untied Transformer++ uses $K$ independently
parameterized blocks.  We also compare with full-attention GAU and with
FLASH~\citep{hua2022transformer}.  All models in
the primary comparison have the same number of distinct non-embedding
parameters, with reused weights counted once, and \cref{app:parameterization}
lists every dimension and realized parameter count.

\section{Experimental Setup}
\label{sec:experiments}

\paragraph{Design and model selection.}
We train full-prefix TANGO, Windowed TANGO, Loop Transformer++, Untied
Transformer++, full-attention GAU, and FLASH at about 70M and 160M
parameters, with depth $K=4$ or $K=8$.  Every combination of dataset, size,
depth, and architecture is trained with three paired model and data seeds,
which gives $4\times2\times2\times6\times3=288$ runs.  Runs that share a data
seed use the same examples in the same order, and every run in a given
dataset and size trains on the same total number of tokens.  During training we
measure development loss at update counts fixed in advance.  For each run, the
checkpoint with the lowest development NLL is selected, and only that
checkpoint is evaluated on the held-out test set, so test data never influence
the choice of checkpoint.  Every table compares all of its models on the same
datasets, token budgets, and seeds.  We also trained every architecture at
depth 16.  We restrict the depth-16 comparison to development NLL and report it
separately from the held-out comparison.

\paragraph{Datasets.}
FineWeb-Edu is a corpus of educational web text~\citep{penedo2024fineweb}.  We
split it by Common Crawl date, tokenize it with the GPT-2 tokenizer, and train
on sequences of 8,192 tokens.  The token budgets are 1.50 billion at 70M and
3.20 billion at 160M.  Lean is an interactive theorem prover and programming
language, and its Mathlib library contains a large body of community-formalized
mathematics~\citep{demoura2021lean4,mathlib2020lean}.  Our Lean task gives
equal weight to modeling Mathlib source files and to completing proofs, where
each proof-completion example conditions on the surrounding context and the
declaration and is scored on the complete proof.  Files are
split so that no training file imports a held-out file, and contexts contain
2,048 tokens.  The token budgets are 1.00 billion and 3.20 billion,
and because the Lean corpus is small, both budgets repeat it many times.
DeepMind Mathematics, which we call mathematics for short, contains
generated problems from 56 modules at three difficulty
levels~\citep{saxton2019analytic}.  We use a character vocabulary
built from the training data.  The example budgets are 5.376 million at 70M
and 12.50 million at 160M.  The original development split of this benchmark
shares about 7\% of its question and answer pairs with the training data, so
we use it only to select checkpoints and report results on the official test
sets, with test NLL measured on the interpolation split.
The Stack v3 provides source code~\citep{lozhkov2026stackv3}, from which we
use six languages in sequences of 8,192 tokens.  Forks of the same repository
are grouped into one family, and each family appears in only one split.  The
token budgets are again 1.50 billion and 3.20 billion.

\paragraph{Models and optimization.}
The names 70M and 160M refer to total parameter counts with the GPT-2
vocabulary.  The 70M models use residual width 512 with 16 heads, and the 160M
models use width 768 with 24 heads, so every head has dimension 32.  We fix the
number of non-embedding parameters at 44.27 million and 121.40 million and then
solve for each architecture's expansion width, which is the gated width $F$
for TANGO and the feed-forward width for the Transformer++ baselines.  Every
realized count is within 0.20\% of its target, and mathematics models use the
same non-embedding targets with a 73-symbol vocabulary.  All models minimize
next-token NLL with AdamW, bfloat16 computation, and 32-bit optimizer states.
The learning rate peaks at $6\times10^{-4}$ and follows a cosine decay, with
weight decay 0.1, gradient clipping at 1.0, and no dropout.  This recipe is
shared by every architecture rather than tuned separately for each.
\Cref{app:experimental-details} gives the full schedules, splits, seeds, and
parameter counts.

\paragraph{TANGO with three gated-projection sets.}
We train TANGO with three gated-projection sets on FineWeb-Edu with the same data, objective, token budgets, and
selection rule as the corresponding models in the primary comparison.  These
models are compared on development data only, so their results complement the
held-out comparison rather than extend it.  Throughput is the number of
training tokens processed per second, taken as the median over each run's final
100 updates and averaged over seeds.  Every run uses an effective batch of 32
sequences, but each model uses its own memory-safe microbatch size with
gradient accumulation (\cref{app:experimental-details}).  The throughput we
report therefore reflects each model's own memory configuration rather than
kernel speed at a shared microbatch size.

\section{Results}
\label{sec:results}

\subsection{Held-out comparison}

\Cref{tab:multiscale-results} contains the central result of the paper.  It
reports test NLL for all \SelectedTestCompleteRuns{} runs, averaged over the
three seeds of each
combination of dataset, size, and depth.

\begin{table*}[h]
 \centering
 \caption{Test NLL at the checkpoint selected on development data, for models
 matched in parameters and training data.  Each entry is a mean over three
 seeds, and lower is better.  Bold marks the lowest mean in each row.  Tier is
 the model size, and depth is the number of block applications.  Stack denotes
 The Stack v3 and DM Math denotes DeepMind Mathematics, where the official
 interpolation test is used.  Lean gives equal weight to source files and proof
 completion.  Values are comparable within a row but not across rows, and
 \cref{tab:full-selected-test} gives standard deviations.}
 \label{tab:multiscale-results}
 \small
 \resizebox{\textwidth}{!}{
\begin{tabular}{lllrrrrrr}
\toprule
Dataset & Tier & Depth & TANGO & \shortstack{Windowed\\TANGO} & \shortstack{Loop\\Transformer++} & \shortstack{Untied\\Transformer++} & \shortstack{Full-attention\\GAU} & FLASH \\
\midrule
FineWeb-Edu & 70M & 4 & \textbf{3.270} & 3.323 & 3.363 & 3.291 & 3.449 & 3.431 \\
FineWeb-Edu & 70M & 8 & \textbf{3.189} & 3.249 & 3.296 & 3.250 & 3.290 & 3.309 \\
FineWeb-Edu & 160M & 4 & \textbf{2.950} & 2.997 & 3.012 & 2.964 & 3.032 & 3.067 \\
FineWeb-Edu & 160M & 8 & \textbf{2.847} & 2.897 & 2.907 & 2.905 & 2.907 & 2.962 \\
Lean & 70M & 4 & \textbf{1.413} & 1.875 & 2.143 & 1.830 & 1.902 & 1.929 \\
Lean & 70M & 8 & \textbf{1.319} & 1.865 & 1.992 & 1.546 & 1.926 & 1.895 \\
Lean & 160M & 4 & \textbf{1.463} & 1.897 & 2.249 & 1.535 & 2.007 & 1.962 \\
Lean & 160M & 8 & \textbf{1.355} & 1.876 & 2.106 & 1.455 & 1.968 & 1.930 \\
DM Math & 70M & 4 & \textbf{0.554} & 0.567 & 0.619 & 0.637 & 0.671 & 0.643 \\
DM Math & 70M & 8 & \textbf{0.463} & 0.472 & 0.594 & 0.607 & 0.551 & 0.544 \\
DM Math & 160M & 4 & \textbf{0.460} & 0.465 & 0.552 & 0.830 & 0.580 & 0.522 \\
DM Math & 160M & 8 & \textbf{0.340} & 0.341 & 0.549 & 0.896 & 0.427 & 0.444 \\
Stack & 70M & 4 & \textbf{0.909} & 1.384 & 1.174 & 0.960 & 1.267 & 1.405 \\
Stack & 70M & 8 & \textbf{0.842} & 1.334 & 1.081 & 0.887 & 1.091 & 1.189 \\
Stack & 160M & 4 & \textbf{0.755} & 1.144 & 1.051 & 0.783 & 0.921 & 1.067 \\
Stack & 160M & 8 & \textbf{0.688} & 1.044 & 1.258 & 0.729 & 0.822 & 0.936 \\
\bottomrule
\end{tabular}
}
\end{table*}

Full-prefix TANGO has the lowest mean test NLL in all 16 comparisons, which
cover web text, Lean source files and proofs, generated mathematics, and code in six
languages.  Its margin over the best Transformer, GAU, or FLASH baseline ranges
from about 0.01 NLL on FineWeb-Edu at 160M and depth 4 to about 0.42 NLL on
Lean at 70M and depth 4.  TANGO is also lower in 47 of the 48 paired-seed
comparisons.  The one
exception is a seed of the
160M, depth-8 mathematics setting, in which Windowed TANGO is lower by 0.0031 NLL.
\Cref{tab:paired-seed-test-margins} gives the margin of every seed against the
lowest-NLL alternative.  A matched single-seed comparison at 410 million
parameters appears in \cref{app:preliminary-410m}.

The ranking of the other architectures depends on the setting.  Untied
Transformer++ ranks second in 10 of the 12 FineWeb-Edu, Lean, and Stack
settings, yet it ranks last in three of the four mathematics settings.  The
lowest-NLL non-TANGO baseline on mathematics is Loop Transformer++, FLASH, or
GAU, depending on size and depth.  Windowed TANGO ranks second in six settings
and last in two.  Across
these comparisons, TANGO ranks first throughout, while the ranks of the other
architectures vary.

At depth 16, TANGO has the
lowest development NLL in seven of the eight settings, and in the eighth,
mathematics at 160M, Windowed TANGO is lowest and TANGO second.  TANGO's
development loss also falls as depth increases from 4 to 8 to 16 in seven of
the eight settings and is essentially unchanged from depth 8 to 16 on Lean at
160M.  The NLL of Loop Transformer++ increases with depth in four of the eight
settings, and that of Untied Transformer++ increases in three.  GAU is the only
other architecture whose loss does not rise as depth increases from 4 to 8 to
16.

On the test sets, the largest standard deviation over seeds is 0.045 NLL for
TANGO and 0.018 for
Windowed TANGO, compared with 0.053 to 0.067 for FLASH, GAU, and Untied
Transformer++ and 0.319 for Loop Transformer++ on The Stack at 160M and depth
8.  At depth 16, on development data, TANGO's largest standard deviation is
0.017, the smallest of any architecture, compared with 0.149 for Loop
Transformer++ and 0.144 for Untied Transformer++.

We also measure exact match on the official mathematics tests.  Exact match is
the fraction of problems whose greedily
decoded answer matches the reference exactly.  At 160M and depth 8, TANGO and
Windowed TANGO answer $51.86\pm0.37\%$ and $52.07\pm0.71\%$ of interpolation
problems exactly, compared with 43.0\% for GAU, the highest value among the
other architectures.
On the extrapolation split, which tests generalization beyond the training
distribution, the corresponding figures are $40.96\pm0.14\%$, $39.72\pm1.68\%$,
and 30.2\%.  Here and below, uncertainties are standard deviations across the
three seeds.  These accuracies remain well below published results that used
far more training examples, as \cref{app:dm-diagnostics} discusses.  The margins
over GAU are about nine percentage points on interpolation and eleven on
extrapolation.

Our primary Untied Transformer++ uses the same residual width as the other
models, which gives it wider feed-forward layers than is conventional, so we
also trained a conventionally proportioned version with a feed-forward ratio near $8/3$
(\cref{app:canonical-untied}).  Across the 24 settings, each with three seeds
for all three models, TANGO has lower development NLL than both Untied
baselines in 22.  The two exceptions are Lean at depth 4.
On mathematics, both Untied
configurations have higher development NLL at 160M than at 70M under the shared
optimization recipe.

The difference between full-prefix and Windowed TANGO is largest on Lean and
code.  At 160M and depth 8, for example, full-prefix TANGO reaches 1.355 on
Lean and 0.688 on The Stack, whereas Windowed TANGO reaches 1.876 and 1.044.
On FineWeb-Edu the two differ by 0.050.  These experiments do not determine
whether the difference comes from access to distant gates, the weighting rule
for older sources, or another architectural difference.

All of these comparisons match parameters and training data, not
computation.  For one 8,192-token FineWeb-Edu sequence at 70M and depth 4,
full-prefix TANGO requires 9.44 trillion analytical forward multiply-accumulate
operations (MACs), compared with 1.94 trillion for Loop Transformer++ and 0.85 trillion for
Untied Transformer++.  Most of this cost comes from averaging the wide gate over
every pair of tokens.  Windowed TANGO avoids that pairwise cost and requires
1.90 trillion MACs at this shape, compared with 1.94 trillion for Loop
Transformer++.  Across the 16 settings, which are matched in parameters and
training data rather than computation, Windowed TANGO has lower test NLL than
Loop Transformer++ in 13.  The cost of Windowed TANGO is linear in sequence
length, whereas the attention cost of Loop Transformer++ is quadratic.
Windowed TANGO also has lower NLL than FLASH, the other linear-cost model, in
13 of the 16 settings.  The three exceptions in both comparisons are on The
Stack.

\subsection{TANGO with three gated-projection sets}
\label{sec:three-set-tango}

\Cref{tab:three-set-fineweb} compares the three-set TANGO variant with TANGO
and the two Transformer++ baselines on FineWeb-Edu.  The table reports
development NLL together with measured training throughput.

\begin{table*}[h]
 \caption{TANGO with three gated-projection sets on FineWeb-Edu.  NLL is the mean $\pm$
 standard deviation over three seeds on development data at the selected
 checkpoint, so these values differ from the test values in
 \cref{tab:multiscale-results}.  Throughput is the mean training rate at the
 end of training, and the final columns divide the three-set model's throughput
 by that of each comparison model.}
 \label{tab:three-set-fineweb}
 \centering
 \small
 \resizebox{\textwidth}{!}{
\begin{tabular}{llrrrrrrrr}
\toprule
Tier & Depth & TANGO NLL & \shortstack{TANGO (3 sets)\\NLL} & Loop NLL & Untied NLL & \shortstack{TANGO (3 sets)\\tok/s} & vs. TANGO & vs. Loop & vs. Untied \\
\midrule
70M & 8 & 3.212 $\pm$ 0.016 & \textbf{3.183 $\pm$ 0.005} & 3.325 $\pm$ 0.007 & 3.277 $\pm$ 0.010 & 48.4k & 3.26$\times$ & 0.96$\times$ & 0.75$\times$ \\
70M & 16 & 3.181 $\pm$ 0.010 & \textbf{3.176 $\pm$ 0.009} & 3.277 $\pm$ 0.001 & 3.296 $\pm$ 0.010 & 26.6k & 3.38$\times$ & 0.75$\times$ & 0.47$\times$ \\
160M & 8 & 2.870 $\pm$ 0.006 & \textbf{2.857 $\pm$ 0.010} & 2.935 $\pm$ 0.002 & 2.930 $\pm$ 0.003 & 27.6k & 3.40$\times$ & 0.91$\times$ & 0.66$\times$ \\
160M & 16 & 2.815 $\pm$ 0.013 & \textbf{2.838 $\pm$ 0.007} & 2.901 $\pm$ 0.005 & 2.937 $\pm$ 0.008 & 14.6k & 3.45$\times$ & 0.79$\times$ & 0.37$\times$ \\
\bottomrule
\end{tabular}
}
\end{table*}

Across the four settings, the three-set model trains 3.26 to 3.45 times faster
than TANGO.  Its mean development NLL is lower than TANGO by 0.005 to 0.029 in
three settings.  At 160M and depth 16, it is higher by 0.023.  The three-set
model has lower NLL than both Transformer++ baselines in all four settings.
It reaches 75\% to 96\% of the throughput of Loop Transformer++ and 37\% to
75\% of the throughput of Untied Transformer++.

Using three gated-projection sets changes both the pattern of parameter sharing
and the gated width that is active at each application.  In a sweep that
holds stored parameters near 70 million and approximately matches the ratio
$F/d$ of gated width to residual width (\cref{app:active-ratio}), the
three-set model is faster at all seven ratios from $0.5$ to $32$ at
depth 4 and has lower NLL at the six ratios up to $16$.  TANGO is lower
by 0.016 NLL at ratio $32$.

\section{Related Work}
\label{sec:related}

\paragraph{Gated token mixing.}
GAU is an architectural antecedent of TANGO.  It replaces attention
and the feed-forward network with one multiplicative residual
update~\citep{hua2022transformer}.  Both models multiply an attention-weighted
aggregate of nonlinear source features by destination-dependent features.  GAU
uses a nonlinear destination branch.  TANGO uses a linear destination branch,
so its source aggregate defines the diagonal core of a linear map on the
destination features.  TANGO also computes attention weights from normalized multihead
cosine scores with a null option.  GAU instead uses a single narrow query and
key projection with squared rectified-linear scores.  MEGA combines a similar single-head gated
attention with an exponential moving average~\citep{ma2023mega}, and gated
softmax attention applies a learned gate to the output of standard value
aggregation~\citep{qiu2025gatedattention}.  Recurrent models such as RWKV and
Gated Linear Attention also multiply an aggregated summary of the prefix by a
gate computed at the destination~\citep{peng2023rwkv,yang2024gated}.  In TANGO,
the source branch is nonlinear and the destination branch is linear.  Our
experiments compare complete architectures and do not isolate the effect of
any one of these differences.

\paragraph{Fast weights and multiplicative conditioning.}
TANGO belongs to a long line of models in which one part of a network produces
the weights or modulations that another part applies.  Fast-weight memories
write weight changes that are later used to transform
inputs~\citep{schmidhuber1992learning}, and linear attention can itself be read
as a fast-weight memory that writes outer products of keys and values and reads
them additively~\citep{schlag2021linear}.  Hypernetworks generate the weights of
a second network~\citep{ha2017hypernetworks}, and LambdaNetworks summarize the
context into linear functions that are applied to each
query~\citep{bello2021lambda}.  Feature-wise linear modulation conditions a
network by scaling and shifting its features~\citep{perez2018film}, and
multiplicative interactions of this kind enlarge the set of functions that a
layer can represent compactly~\citep{jayakumar2020multiplicative}.  TANGO
combines these ideas in a specific form.  Each source gate defines a diagonal
operator, and the attention weights form a mixture of these operators for the
destination features.  This update replaces
both attention and the feed-forward network.

\paragraph{Recurrent depth and parameter sharing.}
Universal Transformers, ALBERT, and deep equilibrium models reuse parameters
across depth~\citep{dehghani2019universal,lan2020albert,bai2019deep}.  Recent
work studies shared blocks at larger scale, sometimes with
iteration encodings, a variable number of iterations, or depth conditioned on
a compute
budget~\citep{saunshi2025reasoning,geiping2025scaling,yang2026loopcoder,jeddi2026loopformer}.
Our primary TANGO models and Loop Transformer++ use a fixed application depth.
Other work shares parameters only partially, either
by assigning different parameters to different layers in fixed
patterns~\citep{takase2023lessons}, by adding layer-specific low-rank
corrections to a shared block~\citep{bae2025relaxed}, or by routing among
shared experts~\citep{csordas2024moeut}.  The recurrent-depth model of
\citet{geiping2025scaling} likewise surrounds a looped core with separate input
and output blocks.  TANGO with three gated-projection sets uses distinct
initial, middle, and final gated projections while sharing the remaining block
parameters across applications.

\paragraph{Efficient attention.}
Windowed TANGO uses the same broad structure as hybrids that pair local
attention with a linear summary of the distant prefix, such as
FLASH~\citep{hua2022transformer} and
BASED~\citep{arora2024simple}.  It uses the same associative running sums, but
the quantities it accumulates are nonlinear source gates that then modulate the
linear features of the destination.

\section{Discussion and Conclusion}
\label{sec:discussion}

Across four domains, two sizes, and two depths, TANGO has lower held-out NLL
than every alternative in the primary comparison when parameters and training
data are matched.  The ordering of the other architectures changes across
settings.  These comparisons concern complete
architectures.  They do not isolate the contributions of the
nonlinear source gate, the linear destination branch, cosine attention, the null
option, or parameter sharing.  The result also does not show a computational
advantage for the full-prefix model, because full-prefix TANGO averages a very
wide gate over every pair of tokens.  Across the four evaluated settings, the
models with three gated-projection sets reach 75\% to 96\% of Loop
Transformer++ throughput.  Their mean development NLL is lower than TANGO in
three settings and higher in one.  These comparisons require held-out
evaluation before they support a conclusion about generalization.

Because one block is shared across applications, each added application adds
computation without adding
parameters.  TANGO's development loss falls as depth increases from 4 to 8 to
16 in seven of the eight settings and rises in none.  Over the same depths,
the development loss of Loop Transformer++, which shares its block in the same
way, rises in four of the eight settings.  These comparisons do not determine
whether TANGO's additional
computation is more useful than spending the same computation in another
architecture.  Such a comparison would hold parameters and data fixed while
matching computation.  At the measured 70M, depth-4 FineWeb-Edu shape,
Windowed TANGO and Loop Transformer++ require similar analytical forward MACs.
Across the 16 settings matched in data and parameters, Windowed TANGO has lower
test NLL in 13.  A looped
Transformer with more applications would provide the corresponding comparison
for full-prefix TANGO.

Whether the learned source gates have consistent semantic roles across
contexts, and whether those gates receive high weight at relevant
destinations, remains to be tested.

\subsection{Limitations}

Three seeds give only coarse estimates of uncertainty, and all architectures
share one optimization recipe, which may favor some families.  Tuning each
architecture separately could therefore improve the baselines whose results
vary substantially across seeds.  Our measurements also leave generation
largely untested, since
likelihood on Lean does not measure whether generated proofs are accepted by
the Lean kernel, and likelihood on code does not measure whether generated
programs run correctly.  Finally, the main comparison matches training data
rather than total computation or wall-clock time.

\subsection{Conclusion}

TANGO replaces separate attention and feed-forward sublayers with a
source-conditioned linear operator whose middle factor is diagonal.  It has
the lowest mean held-out loss in all 16 matched-data, matched-parameter
settings, and its mean development NLL does not increase from depth 4 to 8 to
16 in any setting.  The evaluated three-set models train about
3.3 times faster than TANGO, although they remain slower than
both Transformer++ baselines.

\subsection*{Reproducibility statement}
\Cref{sec:method} specifies the TANGO architecture and the variant with three
gated-projection sets, including every equation and the parameterization.
\Cref{sec:experiments} describes the datasets, training budgets, seeds,
checkpoint selection, and evaluation, and \cref{app:parameterization} gives
every model dimension, optimization schedule, and data split.  We will release
the training and evaluation code, experiment configurations, per-run results,
and table-generation scripts in a public repository linked in the final
version of the paper.

\bibliographystyle{plainnat}
\bibliography{refs}

\appendix
\section*{Overview of the appendices}

\Cref{app:parameterization} gives model dimensions, training details, and
data splits for every experiment, and \cref{app:depth16} reports development
results for every architecture at depth 16.  \Cref{app:canonical-untied} reports an
additional, conventionally proportioned Untied Transformer++ baseline, and
\cref{app:active-width} describes an experiment in which a single TANGO
model is trained so that it can run at several widths.  \Cref{app:three-set-tango}
collects further results for TANGO with three gated-projection sets, and
\cref{app:preliminary-410m} reports a single-seed comparison at 410 million
parameters.  The remaining appendices contain the complete test and
development tables (\cref{app:test-coverage}), the full evaluation on
DeepMind Mathematics (\cref{app:dm-diagnostics}), a different approach to
reducing the cost of full-prefix TANGO (\cref{app:compact-controller}), and
results for each programming language in The Stack
(\cref{app:stack-languages}).

\section{Model dimensions and training details}
\label{app:parameterization}
\label{app:experimental-details}

The names 70M and 160M refer to total parameter counts for models with the
GPT-2 vocabulary.  The 70M models use residual width $d=512$ and 16 heads of
width 32, which gives 44,268,416 distinct non-embedding parameters, and the
160M models use $d=768$ and 24 heads, which gives 121,402,624, counting
reused weights once in both cases.  The GPT-2 vocabulary adds 25,731,584 and
38,597,376 parameters to the tied input and output embedding, and DeepMind
Mathematics uses the same non-embedding targets with a 73-symbol vocabulary.
To meet these targets, we solve for each architecture's expansion width
separately, which gives the configurations in \cref{tab:all-parameter-match}.

\begin{table*}[h]
 \centering
 \caption{Parameter-matched configurations for the primary comparison.  Tier
 is the model size and depth is the number of block applications.  Entries are
 gated, feed-forward, or GAU expansion widths, and GAU is given as a range
 because its learned relative-position parameters depend on context length.
 The Loop and Untied columns refer to the corresponding Transformer++
 baselines.}
 \label{tab:all-parameter-match}
 \small
 \resizebox{\textwidth}{!}{%
 \begin{tabular}{llrrrrrrr}
  \toprule
  Tier & Depth & $d$ & TANGO / Windowed & Loop & Untied & GAU & FLASH & Non-embedding range \\
  \midrule
  70M  & 4 & 512 & 28,480 & 28,144 & 6,528  & 7,136--7,152   & 7,152  & 44.238M--44.307M \\
  70M  & 8 & 512 & 28,480 & 28,144 & 2,912  & 3,536--3,552   & 3,552  & 44.180M--44.279M \\
  160M & 4 & 768 & 52,176 & 51,672 & 12,144 & 13,104--13,120 & 13,120 & 121.293M--121.429M \\
  160M & 8 & 768 & 52,176 & 51,672 & 5,568  & 6,528--6,544   & 6,544  & 121.272M--121.543M \\
  \bottomrule
 \end{tabular}}
\end{table*}

The analytical operation counts in \cref{sec:results} treat one
multiply-accumulate operation as one MAC.  For batch size one, sequence length
$T$, vocabulary size $V$, and $K$ applications, the dense full-prefix counts are
\begin{align}
 C_{\mathrm{TANGO}}
 &=K\!\left[T(2d^2+3dF)+T^2(d+F)\right]+TdV,\\
 C_{\mathrm{Transformer++}}
 &=K\!\left[T(4d^2+3dF)+2T^2d\right]+TdV,
\end{align}
and looped and untied Transformer++ use the second formula with their own
widths.  We do not discount the causal mask, because the measured
implementations form dense score and value products.  The count includes the
output vocabulary projection but excludes embedding lookup, elementwise
operations, normalization, masking, residual additions, backward computation,
optimizer work, memory traffic, and hardware utilization.  Windowed TANGO and
FLASH are counted under the same convention, using the padded block
contractions of their implementations.

Development loss is measured with teacher forcing at update counts fixed
before training, and every
model trained on the same dataset at the same size uses the same schedule.
FineWeb-Edu and The Stack are evaluated at updates 381, 954, 1,907, 3,815, and
5,722 at 70M, with updates 7,629, 9,536, 11,444, and 12,207 added at 160M.
DeepMind Mathematics is evaluated every 21,000 updates through update 84,000 at
70M and every 48,828 updates through update 195,312 at 160M.  Lean is evaluated
on a denser schedule, every 382 updates through update 3,438 and at budget
milestones through update 3,814 at 70M, and every 1,220 updates through update
12,200 and at budget milestones through update 12,208 at 160M.

All models are trained with AdamW using $\beta_1=0.9$, $\beta_2=0.95$,
$\epsilon=10^{-8}$, weight decay 0.1, and gradient clipping at 1.0.  The
learning rate warms up linearly to $6\times10^{-4}$ and then follows a cosine
decay to $6\times10^{-5}$.  FineWeb-Edu and The Stack use 200 warmup updates at
both sizes, Lean uses 76 and 200 warmup updates at 70M and 160M, and DeepMind
Mathematics uses 1,680 and 3,906.  The effective batch contains 32 sequences
for FineWeb-Edu and The Stack, 128 for Lean, and 64 for mathematics.
Microbatch sizes vary by architecture so that each model fits in memory, and
gradient accumulation keeps the effective batch fixed.

The three paired model and data seeds are $(17,101)$, $(23,103)$, and
$(41,107)$.  All runs use eager PyTorch 2.8 with bfloat16 computation on one
NVIDIA H100 graphics processor with 80 gigabytes of memory.  The throughput
measurements in \cref{sec:three-set-tango} also use an effective batch of 32 sequences
with a separate microbatch size for each model.  At 70M and depth 8,
TANGO with three gated-projection sets, TANGO, Loop Transformer++, and Untied Transformer++
use microbatches of 4, 8, 16, and 32 sequences with 8, 4, 2, and 1 accumulation
steps.  At 70M and depth 16 they use microbatches of 2, 4, 8, and 32 with 16, 8,
4, and 1 accumulation steps, and at 160M and depth 8 they use microbatches of
2, 8, 8, and 32 with 16, 4, 4, and 1 accumulation steps.  The configurations at
70M and depth 8 are the largest that fit in memory in our capacity tests,
whereas the other settings use more conservative choices.

TANGO and the Transformer++ baselines initialize linear and embedding weights
from $\mathcal N(0,0.02^2)$ and set all biases to zero.  Residual output
projections use standard deviation $0.02/\sqrt{2K}$, and the output projection
of each gated-update set is scaled by an additional factor of $\sqrt{d/F}$.
Attention temperatures are initialized to $\sqrt{d_h}$, where $d_h$ is the attention head
dimension, and null biases are initialized to zero.  The parameter-matched GAU
uses analogous width and depth scaling, while FLASH keeps the initialization of
its reference implementation.  All root mean square layer normalizations use
$\epsilon=10^{-5}$, and rotary position embeddings use base 10,000.  GAU uses
one attention head with query and key dimension 128, and the FLASH reference
uses one local and one global head with chunk size 256.

The FineWeb-Edu training, development, and test sets come from Common Crawl
dumps through 2023-40, 2023-50, and 2024-10, respectively, and we deduplicate
by normalized web address and by exact content across these periods.  The
resulting test set has 4,096 documents and 4.04 million scored tokens.  Lean
uses Mathlib commit \texttt{cf8e23a62939}, split so that no training file
imports a held-out file, and its test set has 225 source segments and 1,554
proof completions.  The official mathematics test sets contain 560,000
interpolation examples and 150,000 extrapolation examples, and greedy decoding
for exact match is capped at 30 answer characters plus the end-of-sequence
symbol.  The original development split of DeepMind Mathematics shares 7.03\%
of its normalized question and answer pairs with the training data, which is
why we use it only to select checkpoints and evaluate on the official test
sets.  The Stack uses Python, Java, JavaScript, C++, Go, and Rust, no
repository family appears in more than one split, and the test set contains
3,072 files from 1,837 repository families.

\section{Development results at depth 16}
\label{app:depth16}

We also trained every architecture at depth 16, with the same data, token
budgets, optimization recipe, and checkpoint-selection rule as the primary
comparison.  The depth-16 models keep the residual widths and approximately the
same non-embedding parameter targets as the shallower models, so the untied
architectures have narrower layers at depth 16.  We restrict this comparison to
development data.  Every value is the best development NLL over a run's
evaluation schedule, which is the quantity used to select checkpoints.
\Cref{tab:depth16-development} reports all seven
architectures, including the conventionally proportioned Untied baseline of
\cref{app:canonical-untied}, and \cref{tab:depth-ordering} follows TANGO from
depth 4 to depth 16.

\begin{table*}[h]
 \centering
 \caption{Best development NLL at depth 16, mean $\pm$ standard deviation over
 three seeds.  Bold marks the lowest mean in each row.  This appendix reports
 development results only.}
 \label{tab:depth16-development}
 \small
 \resizebox{\textwidth}{!}{
\begin{tabular}{llrrrrrrr}
\toprule
Dataset & Tier & TANGO & \shortstack{Windowed\\TANGO} & \shortstack{Loop\\Transformer++} & \shortstack{Untied\\Transformer++} & \shortstack{Conventional\\Untied} & \shortstack{Full-attention\\GAU} & FLASH \\
\midrule
FineWeb-Edu & 70M & \textbf{3.181 $\pm$ 0.010} & 3.259 $\pm$ 0.008 & 3.277 $\pm$ 0.001 & 3.296 $\pm$ 0.010 & 3.293 $\pm$ 0.017 & 3.223 $\pm$ 0.004 & 3.266 $\pm$ 0.009 \\
FineWeb-Edu & 160M & \textbf{2.815 $\pm$ 0.013} & 2.888 $\pm$ 0.005 & 2.901 $\pm$ 0.005 & 2.937 $\pm$ 0.008 & 2.936 $\pm$ 0.006 & 2.857 $\pm$ 0.007 & 2.904 $\pm$ 0.004 \\
Lean & 70M & \textbf{1.477 $\pm$ 0.017} & 2.062 $\pm$ 0.028 & 2.247 $\pm$ 0.048 & 1.718 $\pm$ 0.144 & 1.619 $\pm$ 0.041 & 2.178 $\pm$ 0.023 & 2.179 $\pm$ 0.019 \\
Lean & 160M & \textbf{1.570 $\pm$ 0.015} & 2.086 $\pm$ 0.018 & 2.333 $\pm$ 0.037 & 1.624 $\pm$ 0.013 & 1.592 $\pm$ 0.012 & 2.176 $\pm$ 0.026 & 2.213 $\pm$ 0.027 \\
DM Math & 70M & \textbf{0.379 $\pm$ 0.006} & 0.384 $\pm$ 0.004 & 0.671 $\pm$ 0.144 & 0.511 $\pm$ 0.010 & 0.503 $\pm$ 0.005 & 0.448 $\pm$ 0.004 & 0.429 $\pm$ 0.008 \\
DM Math & 160M & 0.240 $\pm$ 0.002 & \textbf{0.232 $\pm$ 0.004} & 0.753 $\pm$ 0.149 & 0.786 $\pm$ 0.012 & 0.849 $\pm$ 0.027 & 0.318 $\pm$ 0.003 & 0.314 $\pm$ 0.004 \\
Stack & 70M & \textbf{0.802 $\pm$ 0.001} & 1.329 $\pm$ 0.032 & 1.105 $\pm$ 0.062 & 0.866 $\pm$ 0.014 & 0.865 $\pm$ 0.006 & 1.025 $\pm$ 0.009 & 1.132 $\pm$ 0.004 \\
Stack & 160M & \textbf{0.655 $\pm$ 0.002} & 0.985 $\pm$ 0.002 & 1.397 $\pm$ 0.138 & 0.709 $\pm$ 0.007 & 0.710 $\pm$ 0.009 & 0.749 $\pm$ 0.003 & 0.846 $\pm$ 0.037 \\
\bottomrule
\end{tabular}
}
\end{table*}

\begin{table}[h]
 \centering
 \caption{Best development NLL of TANGO at depths 4, 8, and 16.  Depth-4 and
 depth-8 values are the three-seed means of \cref{tab:development-results}, and
 depth-16 values are means $\pm$ standard deviations over three seeds.}
 \label{tab:depth-ordering}
 \small
\begin{tabular}{llrrr}
\toprule
Dataset & Tier & Depth 4 & Depth 8 & Depth 16 \\
\midrule
FineWeb-Edu & 70M & 3.293 & 3.212 & 3.181 $\pm$ 0.010 \\
FineWeb-Edu & 160M & 2.973 & 2.870 & 2.815 $\pm$ 0.013 \\
Lean & 70M & 1.649 & 1.522 & 1.477 $\pm$ 0.017 \\
Lean & 160M & 1.698 & 1.570 & 1.570 $\pm$ 0.015 \\
DM Math & 70M & 0.473 & 0.386 & 0.379 $\pm$ 0.006 \\
DM Math & 160M & 0.383 & 0.270 & 0.240 $\pm$ 0.002 \\
Stack & 70M & 0.899 & 0.832 & 0.802 $\pm$ 0.001 \\
Stack & 160M & 0.747 & 0.680 & 0.655 $\pm$ 0.002 \\
\bottomrule
\end{tabular}

\end{table}

TANGO has the lowest development NLL at depth 16 in seven of the eight
settings.  In mathematics at 160M, Windowed TANGO is lowest at 0.232, TANGO is
second at 0.240, and the best other architecture, FLASH, reaches 0.314.
TANGO's development loss falls from depth 4 to depth 8 to depth 16 in seven of
the eight settings, and on Lean at 160M the means at depths 8 and 16 differ by
less than 0.0001.  Among the other architectures, GAU improves with every added
depth in all eight settings, FLASH and Windowed TANGO in six, Untied
Transformer++ in five, and Loop Transformer++ in four.

\section{Conventionally proportioned Untied Transformer++}
\label{app:canonical-untied}

The primary comparison fixes the residual width at each size and then solves
for each architecture's expansion width.  For Untied Transformer++, this choice
gives SwiGLU expansion ratios $F/d$ of 12.75, 5.69, and 2.80 at 70M with depths
4, 8, and 16, and of 15.81, 7.25, and 2.60 at 160M.  The ratios at depths 4 and
8 exceed the conventional value
of $8/3$, so we also train a baseline whose ratio is close to $8/3$, with the
residual width chosen to keep the total parameter count within 0.5\% of the
same size.  We call this
model the conventionally proportioned Untied baseline, and we report it in
addition to the original baseline rather than in place of it.  Unlike the
primary comparison, this baseline matches total rather than non-embedding
parameters.  A different residual width also changes the size of the tied
embedding, so the division of parameters between the sequence model and the
token interface changes as well.

\begin{table}[h]
 \centering
 \caption{Geometry of the conventionally proportioned Untied baselines at depths
 4, 8, and 16.  The head dimension is 32.  Counts use the GPT-2 vocabulary and count
 the tied input and output embedding once.}
 \label{tab:canonical-untied-geometry}
 \small
 \begin{tabular}{llrrrrrr}
  \toprule
  Tier & Depth & $d$ & Heads & $F$ & $F/d$ & Non-embedding & Total \\
  \midrule
  70M  & 4 & 800   & 25 & 2,064 & 2.580 & 30,061,600  & 70,267,200 \\
  70M  & 8 & 640   & 20 & 1,616 & 2.525 & 37,939,840  & 70,104,320 \\
  70M  & 16 & 480  & 15 & 1,344 & 2.800 & 45,727,200  & 69,850,560 \\
  160M & 4 & 1,376 & 43 & 3,664 & 2.663 & 90,806,368  & 159,960,000 \\
  160M & 8 & 1,056 & 33 & 2,816 & 2.667 & 107,071,008 & 160,142,400 \\
  160M & 16 & 800 & 25 & 2,080 & 2.600 & 120,858,400 & 161,064,000 \\
  \bottomrule
 \end{tabular}
\end{table}

\begin{table*}[h]
 \centering
 \caption{Development NLL at the selected checkpoint in all 24 settings.  Each
 entry is a mean $\pm$ standard deviation over three seeds.  Bold marks the
 lowest mean among the three architectures shown, and the final column
 subtracts the original Untied result from the conventional one.  This
 comparison uses development results only.}
 \label{tab:canonical-untied-validation}
 \small
 \resizebox{\textwidth}{!}{%
\begin{tabular}{lllrrrr}
\toprule
Dataset & Tier & Depth & TANGO & Original-grid Untied & Conventional Untied & $\Delta$(conventional--original) \\
\midrule
FineWeb-Edu & 70M & 4 & \textbf{3.293 $\pm$ 0.005} & 3.318 $\pm$ 0.006 & 3.321 $\pm$ 0.003 & +0.003 \\
FineWeb-Edu & 70M & 8 & \textbf{3.212 $\pm$ 0.016} & 3.277 $\pm$ 0.010 & 3.285 $\pm$ 0.003 & +0.007 \\
FineWeb-Edu & 70M & 16 & \textbf{3.181 $\pm$ 0.010} & 3.296 $\pm$ 0.010 & 3.293 $\pm$ 0.017 & -0.003 \\
FineWeb-Edu & 160M & 4 & \textbf{2.973 $\pm$ 0.008} & 2.990 $\pm$ 0.003 & 3.010 $\pm$ 0.002 & +0.019 \\
FineWeb-Edu & 160M & 8 & \textbf{2.870 $\pm$ 0.006} & 2.930 $\pm$ 0.003 & 2.958 $\pm$ 0.008 & +0.028 \\
FineWeb-Edu & 160M & 16 & \textbf{2.815 $\pm$ 0.013} & 2.937 $\pm$ 0.008 & 2.936 $\pm$ 0.006 & -0.001 \\
Lean & 70M & 4 & 1.649 $\pm$ 0.012 & 2.095 $\pm$ 0.034 & \textbf{1.609 $\pm$ 0.025} & -0.486 \\
Lean & 70M & 8 & \textbf{1.522 $\pm$ 0.017} & 1.784 $\pm$ 0.067 & 1.722 $\pm$ 0.069 & -0.061 \\
Lean & 70M & 16 & \textbf{1.477 $\pm$ 0.017} & 1.718 $\pm$ 0.144 & 1.619 $\pm$ 0.041 & -0.099 \\
Lean & 160M & 4 & 1.698 $\pm$ 0.030 & 1.792 $\pm$ 0.084 & \textbf{1.639 $\pm$ 0.015} & -0.153 \\
Lean & 160M & 8 & \textbf{1.570 $\pm$ 0.016} & 1.687 $\pm$ 0.037 & 1.607 $\pm$ 0.021 & -0.080 \\
Lean & 160M & 16 & \textbf{1.570 $\pm$ 0.015} & 1.624 $\pm$ 0.013 & 1.592 $\pm$ 0.012 & -0.032 \\
DM Math & 70M & 4 & \textbf{0.473 $\pm$ 0.008} & 0.556 $\pm$ 0.005 & 0.786 $\pm$ 0.006 & +0.230 \\
DM Math & 70M & 8 & \textbf{0.386 $\pm$ 0.005} & 0.531 $\pm$ 0.006 & 0.611 $\pm$ 0.007 & +0.080 \\
DM Math & 70M & 16 & \textbf{0.379 $\pm$ 0.006} & 0.511 $\pm$ 0.010 & 0.503 $\pm$ 0.005 & -0.008 \\
DM Math & 160M & 4 & \textbf{0.383 $\pm$ 0.004} & 0.740 $\pm$ 0.034 & 1.171 $\pm$ 0.012 & +0.431 \\
DM Math & 160M & 8 & \textbf{0.270 $\pm$ 0.004} & 0.809 $\pm$ 0.035 & 1.123 $\pm$ 0.018 & +0.314 \\
DM Math & 160M & 16 & \textbf{0.240 $\pm$ 0.002} & 0.786 $\pm$ 0.012 & 0.849 $\pm$ 0.027 & +0.062 \\
Stack & 70M & 4 & \textbf{0.899 $\pm$ 0.003} & 0.948 $\pm$ 0.004 & 0.928 $\pm$ 0.001 & -0.020 \\
Stack & 70M & 8 & \textbf{0.832 $\pm$ 0.003} & 0.876 $\pm$ 0.007 & 0.873 $\pm$ 0.011 & -0.003 \\
Stack & 70M & 16 & \textbf{0.802 $\pm$ 0.001} & 0.866 $\pm$ 0.014 & 0.865 $\pm$ 0.006 & -0.002 \\
Stack & 160M & 4 & \textbf{0.747 $\pm$ 0.005} & 0.774 $\pm$ 0.004 & 0.778 $\pm$ 0.000 & +0.004 \\
Stack & 160M & 8 & \textbf{0.680 $\pm$ 0.002} & 0.721 $\pm$ 0.002 & 0.721 $\pm$ 0.003 & +0.000 \\
Stack & 160M & 16 & \textbf{0.655 $\pm$ 0.002} & 0.709 $\pm$ 0.007 & 0.710 $\pm$ 0.009 & +0.002 \\
\bottomrule
\end{tabular}
}
\end{table*}

The comparison covers all 24 settings with three seeds for every model.  TANGO
has lower development NLL than both Untied baselines in 22.  In the other two,
Lean at 70M and 160M with depth 4, the conventional baseline is lower than
TANGO by 0.040 and 0.059 NLL.  The conventional proportions substantially
improve the Untied baseline on Lean.  Their effect is smaller on FineWeb-Edu
and The Stack, where the change from the original baseline is at most 0.028
NLL, and they increase NLL in five of the six mathematics settings.  The
baseline proportions therefore do not explain TANGO's advantage outside the
two Lean settings at depth 4.

\section{Rotating narrow-width training}
\label{app:active-width}

The three-set TANGO variant lowers cost by narrowing the gated projections
active at each application.  The rotating-width experiment tests whether a single stored
set of parameters can be trained so that each application runs only a small
part of its gated or feed-forward width.  The
rotating model keeps all of the roughly 70M stored parameters.  Its width is
divided into five disjoint segments of equal size, each containing about 20\%
of the real channels.  Each application uses one segment, and the active
segment rotates so that every segment receives the same amount of training.
As a comparison, we also train a separate model whose stored width equals one
active segment.  These narrow models contain 35 to 42 million total
parameters.  For efficient kernels, the active widths are padded, from
356 to 368 channels per attention head for TANGO and from 583 to 640 channels for
Untied Transformer++, while Loop Transformer++ uses 5,632 active channels.

During training, a seeded schedule selects a contiguous segment at each
application and rotates the selection across applications and training
sequences.  At evaluation, a starting offset determines which segment is used
first in the cyclic schedule at depth 8.  We evaluate all five starting segments
at each of the five saved checkpoints and average the development NLL over
starting segments.  We then select the checkpoint with the lowest equal-weight
average of this narrow-execution NLL and the NLL obtained with all segments
active, a selection rule fixed before training.  The separate narrow model uses
its own checkpoint selected on development data.  \Cref{tab:active-width-w0-w4}
reports means and standard deviations over three seeds, and it includes the
ordinary 70M, depth-8 model trained at full width for reference.

\begin{table*}[h]
 \centering
 \caption{FineWeb-Edu development NLL at 70M and depth 8 under rotating
 narrow-width training.  The rotating-sparse and all-segments-active columns run
 the same stored checkpoint, and the narrow specialist is a separately trained
 model.  The final column subtracts the specialist's NLL from the
 rotating-sparse NLL, so negative values favor the rotating model.  Packed
 widths are per attention head for the TANGO variants and total feed-forward
 width for the Transformer++ variants.}
 \label{tab:active-width-w0-w4}
 \small
 \resizebox{\textwidth}{!}{%
\begin{tabular}{lrrrrrr}
\toprule
Architecture & Active packed width & Dense-trained & Rotating sparse & All segments active & Narrow specialist & $\Delta$(sparse--specialist) \\
\midrule
TANGO & 368 & 3.212 $\pm$ 0.016 & 3.397 $\pm$ 0.012 & 3.353 $\pm$ 0.010 & 3.431 $\pm$ 0.005 & -0.034 \\
Windowed TANGO & 368 & 3.282 $\pm$ 0.001 & 3.488 $\pm$ 0.002 & 3.446 $\pm$ 0.002 & 3.507 $\pm$ 0.004 & -0.019 \\
Loop Transformer++ & 5,632 & 3.325 $\pm$ 0.007 & 3.449 $\pm$ 0.004 & 4.223 $\pm$ 0.045 & 3.501 $\pm$ 0.004 & -0.052 \\
Untied Transformer++ & 640 & 3.277 $\pm$ 0.010 & 3.562 $\pm$ 0.012 & 4.214 $\pm$ 0.131 & 3.460 $\pm$ 0.008 & +0.102 \\
\bottomrule
\end{tabular}
}
\end{table*}

Both TANGO variants can run either one segment or all segments of the same
checkpoint, and the difference in loss between the two is small.  Running one segment is 0.043 NLL worse than running
all segments for full-prefix TANGO and 0.042 worse for Windowed TANGO, and it is
still 0.034 and 0.019 NLL better than the corresponding narrow model.  The
Transformer++ baselines have higher NLL when all segments are active than when
one segment is active.  Loop Transformer++ has lower NLL than its narrow model
when running one segment, whereas Untied Transformer++ has higher NLL.  The two
evaluated execution widths therefore remain close in quality for TANGO but not
for the Transformer++ baselines.  For the two TANGO variants, narrow execution
costs 0.185 to 0.206 NLL relative to ordinary training at full width.

These are development results, and they concern quality rather than cost.
Narrow execution removes different operations in different architectures, and
the rotating model still stores the full parameters and optimizer state, so the
table does not show a saving in time or in total memory.

\section{Additional results for TANGO with three gated-projection sets}
\label{app:three-set-tango}

\subsection{Matched active gated-width ratio}
\label{app:active-ratio}

Using three gated-projection sets changes both parameter sharing and active
gated width.  To compare TANGO and the three-set model at similar active widths,
this experiment fixes a
budget of about 70 million stored parameters and approximately matches
the ratio $F/d$ between separately constructed models, where $F$ is the active
gated width and $d$ is the residual width.  Because
the head dimension stays at 32, changing $d$ also changes the number of
attention heads.  \Cref{tab:active-ratio} reports seven ratios at depth 4 with
three paired seeds for each model.  NLL is measured on development data, and
throughput is the rate at the end of training in training tokens per second.

\begin{table*}[h]
 \centering
 \caption{Model dimensions for the active-ratio comparison at depth 4.
 Parameter counts include the tied token embedding once.}
 \label{tab:active-ratio-geometry}
 \small
 \resizebox{\textwidth}{!}{%
\begin{tabular}{rrrrrrrrrrr}
\toprule
Target & \multicolumn{5}{c}{TANGO} & \multicolumn{5}{c}{TANGO (3 sets)} \\
\cmidrule(lr){2-6}\cmidrule(lr){7-11}
$F/d$ & $d$ & Heads & $F$ & Realized & Params & $d$ & Heads & $F$ & Realized & Params \\
\midrule
0.5 & 1,280 & 40 & 640 & 0.500 & 70.066M & 1,216 & 38 & 570 & 0.469 & 70.310M \\
1 & 1,248 & 39 & 1,170 & 0.938 & 70.219M & 1,120 & 35 & 1,120 & 1.000 & 70.089M \\
2 & 1,184 & 37 & 2,257 & 1.906 & 70.327M & 992 & 31 & 2,015 & 2.031 & 69.815M \\
4 & 1,088 & 34 & 4,046 & 3.719 & 70.255M & 864 & 27 & 3,267 & 3.781 & 70.321M \\
8 & 928 & 29 & 7,656 & 8.250 & 69.677M & 704 & 22 & 5,346 & 7.594 & 70.246M \\
16 & 768 & 24 & 12,984 & 16.906 & 69.694M & 544 & 17 & 8,653 & 15.906 & 70.298M \\
32 & 640 & 20 & 19,460 & 30.406 & 70.348M & 416 & 13 & 13,104 & 31.500 & 70.315M \\
\bottomrule
\end{tabular}
}
\end{table*}

\begin{table}[h]
 \centering
 \caption{Active-ratio comparison at 70M and depth 4.  NLL entries and the
 paired differences between the three-set model and TANGO are means $\pm$
 standard deviations over three paired seeds, and negative differences favor
 the three-set model.
 Rates are means over seeds of each run's median over its final 100 updates.}
 \label{tab:active-ratio}
 \small
 \resizebox{\columnwidth}{!}{%
\begin{tabular}{rrrrrr}
\toprule
Target $F/d$ & TANGO NLL & \shortstack{TANGO (3 sets)\\NLL} & $\Delta$ NLL & TANGO tok/s & \shortstack{TANGO (3 sets)\\tok/s} \\
\midrule
0.5 & 3.744 $\pm$ 0.009 & 3.575 $\pm$ 0.009 & -0.169 $\pm$ 0.011 & 82.2k & 95.1k \\
1 & 3.591 $\pm$ 0.009 & 3.446 $\pm$ 0.009 & -0.145 $\pm$ 0.018 & 80.4k & 103.9k \\
2 & 3.473 $\pm$ 0.005 & 3.363 $\pm$ 0.009 & -0.111 $\pm$ 0.014 & 72.3k & 95.7k \\
4 & 3.384 $\pm$ 0.015 & 3.305 $\pm$ 0.006 & -0.079 $\pm$ 0.010 & 69.7k & 95.9k \\
8 & 3.315 $\pm$ 0.013 & 3.277 $\pm$ 0.007 & -0.037 $\pm$ 0.016 & 59.3k & 90.9k \\
16 & 3.281 $\pm$ 0.014 & 3.271 $\pm$ 0.013 & -0.009 $\pm$ 0.020 & 47.5k & 73.0k \\
32 & 3.271 $\pm$ 0.008 & 3.287 $\pm$ 0.002 & +0.016 $\pm$ 0.010 & 33.7k & 70.8k \\
\bottomrule
\end{tabular}
}
\end{table}

The three-set model has lower NLL at the six ratios from $0.5$ to $16$ and is
faster at all seven ratios.  At ratio $32$, TANGO has 0.016 lower NLL, while
the three-set model still processes 2.10 times as many training tokens per
second.  At a matched stored budget and an approximately matched active ratio,
the evaluated three-set configurations have higher throughput at every
reported ratio.  Their NLL is
lower at six of the seven ratios.  Because residual width and head count vary together
with the ratio, however, the sweep does not isolate the effect of using
distinct gated projections for the first, middle, and final applications.

\section{Single-seed comparison at 410M parameters}
\label{app:preliminary-410m}

We report a matched single-seed comparison at 410 million parameters on
FineWeb-Edu.  The comparison uses development measurements for seed pair
$(17,101)$ after 3.20 billion training tokens at depth 8.  We restrict this
comparison to development data.  All three models use residual width 1,024 and 32
heads.  TANGO and Windowed TANGO use gated width 116,032 and have 358,549,568
non-embedding parameters, and Untied Transformer++ uses SwiGLU width 13,216 and
has 358,368,256 non-embedding parameters.

\begin{table}[h]
 \centering
 \caption{FineWeb-Edu development NLL at 410M and depth 8 after the
 same number of training tokens.  Every entry uses seed pair $(17,101)$ after
 3.20 billion tokens.}
 \label{tab:preliminary-410m}
 \small
 \resizebox{\textwidth}{!}{
\begin{tabular}{llrrrr}
\toprule
Presented tokens & Seed pair & TANGO & Windowed TANGO & Untied & $\Delta$(TANGO--Untied) \\
\midrule
3.20 billion & 17/101 & 2.745 & 2.823 & 2.835 & -0.090 \\
\bottomrule
\end{tabular}
}
\end{table}

In this seed, TANGO has lower development NLL than both Windowed TANGO and
Untied Transformer++.  A single seed at one point in training does not
establish a trend with scale.

\section{Complete test and development tables}
\label{app:test-coverage}

The primary held-out comparison contains all 288 runs.  Each of the 16
combinations of dataset, size, and depth includes all six architectures and
three paired seeds.  \Cref{tab:full-selected-test} adds standard deviations to the test
results of \cref{tab:multiscale-results}, \cref{tab:paired-seed-test-margins}
gives TANGO's margin in each individual seed, and
\cref{tab:development-results} reports the development results that were used
to select checkpoints.

\begin{table*}[ht]
  \centering
  \caption{Test NLL at the selected checkpoint, mean $\pm$ standard deviation
  over three seeds.  Mathematics uses the official interpolation test, and
  bold marks the lowest mean in each row.}
  \label{tab:full-selected-test}
  \resizebox{\textwidth}{!}{%
\begin{tabular}{lllrrrrrr}
\toprule
Dataset & Tier & Depth & TANGO & \shortstack{Windowed\\TANGO} & \shortstack{Loop\\Transformer++} & \shortstack{Untied\\Transformer++} & \shortstack{Full-attention\\GAU} & FLASH \\
\midrule
FineWeb-Edu & 70M & 4 & \textbf{3.270 $\pm$ 0.005} & 3.323 $\pm$ 0.005 & 3.363 $\pm$ 0.003 & 3.291 $\pm$ 0.006 & 3.449 $\pm$ 0.004 & 3.431 $\pm$ 0.020 \\
FineWeb-Edu & 70M & 8 & \textbf{3.189 $\pm$ 0.016} & 3.249 $\pm$ 0.001 & 3.296 $\pm$ 0.006 & 3.250 $\pm$ 0.011 & 3.290 $\pm$ 0.019 & 3.309 $\pm$ 0.006 \\
FineWeb-Edu & 160M & 4 & \textbf{2.950 $\pm$ 0.008} & 2.997 $\pm$ 0.002 & 3.012 $\pm$ 0.007 & 2.964 $\pm$ 0.003 & 3.032 $\pm$ 0.013 & 3.067 $\pm$ 0.019 \\
FineWeb-Edu & 160M & 8 & \textbf{2.847 $\pm$ 0.006} & 2.897 $\pm$ 0.004 & 2.907 $\pm$ 0.001 & 2.905 $\pm$ 0.004 & 2.907 $\pm$ 0.012 & 2.962 $\pm$ 0.015 \\
Lean & 70M & 4 & \textbf{1.413 $\pm$ 0.011} & 1.875 $\pm$ 0.016 & 2.143 $\pm$ 0.030 & 1.830 $\pm$ 0.027 & 1.902 $\pm$ 0.051 & 1.929 $\pm$ 0.046 \\
Lean & 70M & 8 & \textbf{1.319 $\pm$ 0.021} & 1.865 $\pm$ 0.013 & 1.992 $\pm$ 0.021 & 1.546 $\pm$ 0.044 & 1.926 $\pm$ 0.034 & 1.895 $\pm$ 0.032 \\
Lean & 160M & 4 & \textbf{1.463 $\pm$ 0.045} & 1.897 $\pm$ 0.006 & 2.249 $\pm$ 0.019 & 1.535 $\pm$ 0.066 & 2.007 $\pm$ 0.028 & 1.962 $\pm$ 0.050 \\
Lean & 160M & 8 & \textbf{1.355 $\pm$ 0.026} & 1.876 $\pm$ 0.018 & 2.106 $\pm$ 0.018 & 1.455 $\pm$ 0.031 & 1.968 $\pm$ 0.047 & 1.930 $\pm$ 0.053 \\
DM Math & 70M & 4 & \textbf{0.554 $\pm$ 0.004} & 0.567 $\pm$ 0.002 & 0.619 $\pm$ 0.004 & 0.637 $\pm$ 0.002 & 0.671 $\pm$ 0.006 & 0.643 $\pm$ 0.009 \\
DM Math & 70M & 8 & \textbf{0.463 $\pm$ 0.009} & 0.472 $\pm$ 0.004 & 0.594 $\pm$ 0.001 & 0.607 $\pm$ 0.008 & 0.551 $\pm$ 0.006 & 0.544 $\pm$ 0.014 \\
DM Math & 160M & 4 & \textbf{0.460 $\pm$ 0.002} & 0.465 $\pm$ 0.003 & 0.552 $\pm$ 0.011 & 0.830 $\pm$ 0.031 & 0.580 $\pm$ 0.020 & 0.522 $\pm$ 0.018 \\
DM Math & 160M & 8 & \textbf{0.340 $\pm$ 0.003} & 0.341 $\pm$ 0.004 & 0.549 $\pm$ 0.007 & 0.896 $\pm$ 0.039 & 0.427 $\pm$ 0.006 & 0.444 $\pm$ 0.014 \\
Stack & 70M & 4 & \textbf{0.909 $\pm$ 0.004} & 1.384 $\pm$ 0.010 & 1.174 $\pm$ 0.018 & 0.960 $\pm$ 0.004 & 1.267 $\pm$ 0.067 & 1.405 $\pm$ 0.051 \\
Stack & 70M & 8 & \textbf{0.842 $\pm$ 0.003} & 1.334 $\pm$ 0.018 & 1.081 $\pm$ 0.013 & 0.887 $\pm$ 0.007 & 1.091 $\pm$ 0.015 & 1.189 $\pm$ 0.016 \\
Stack & 160M & 4 & \textbf{0.755 $\pm$ 0.005} & 1.144 $\pm$ 0.001 & 1.051 $\pm$ 0.008 & 0.783 $\pm$ 0.004 & 0.921 $\pm$ 0.012 & 1.067 $\pm$ 0.050 \\
Stack & 160M & 8 & \textbf{0.688 $\pm$ 0.002} & 1.044 $\pm$ 0.009 & 1.258 $\pm$ 0.319 & 0.729 $\pm$ 0.002 & 0.822 $\pm$ 0.012 & 0.936 $\pm$ 0.047 \\
\bottomrule
\end{tabular}
  }
\end{table*}

\begin{table}[ht]
  \centering
  \caption{Test NLL margin of TANGO over the best other model in the primary
  comparison, for each seed.  Negative values favor TANGO.  Mathematics uses
  the official interpolation test, and each value compares the six
  architectures at the same model and data-order seed.}
  \label{tab:paired-seed-test-margins}
  \small
  \resizebox{\columnwidth}{!}{%
\begin{tabular}{lllrrr}
\toprule
Dataset & Tier & Depth & Seed 17/101 & Seed 23/103 & Seed 41/107 \\
\midrule
FineWeb-Edu & 70M & 4 & -0.0282 & -0.0219 & -0.0132 \\
FineWeb-Edu & 70M & 8 & -0.0338 & -0.0761 & -0.0588 \\
FineWeb-Edu & 160M & 4 & -0.0112 & -0.0252 & -0.0062 \\
FineWeb-Edu & 160M & 8 & -0.0444 & -0.0608 & -0.0455 \\
Lean & 70M & 4 & -0.4217 & -0.4235 & -0.3939 \\
Lean & 70M & 8 & -0.2409 & -0.2840 & -0.1566 \\
Lean & 160M & 4 & -0.0419 & -0.1119 & -0.0602 \\
Lean & 160M & 8 & -0.1252 & -0.1375 & -0.0383 \\
DM Math & 70M & 4 & -0.0105 & -0.0169 & -0.0113 \\
DM Math & 70M & 8 & -0.0036 & -0.0134 & -0.0099 \\
DM Math & 160M & 4 & -0.0061 & -0.0050 & -0.0034 \\
DM Math & 160M & 8 & -0.0039 & +0.0031 & -0.0009 \\
Stack & 70M & 4 & -0.0428 & -0.0551 & -0.0546 \\
Stack & 70M & 8 & -0.0411 & -0.0480 & -0.0455 \\
Stack & 160M & 4 & -0.0325 & -0.0309 & -0.0200 \\
Stack & 160M & 8 & -0.0424 & -0.0403 & -0.0399 \\
\bottomrule
\end{tabular}
  }
\end{table}

\begin{table*}[ht]
  \centering
  \caption{Development NLL at the checkpoint selected from each run's
  evaluation schedule, mean $\pm$ standard deviation over three seeds.  These
  values cover all 288 training runs and were used to select checkpoints, so
  they are separate from the test results.}
  \label{tab:development-results}
  \resizebox{\textwidth}{!}{%
\begin{tabular}{lllrrrrrr}
\toprule
Dataset & Tier & Depth & TANGO & \shortstack{Windowed\\TANGO} &
\shortstack{Loop\\Transformer++} & \shortstack{Untied\\Transformer++} &
\shortstack{Full-attention\\GAU} & FLASH \\
\midrule
FineWeb-Edu & 70M  & 4 & \textbf{3.293 $\pm$ 0.005} & 3.355 $\pm$ 0.005 & 3.392 $\pm$ 0.003 & 3.318 $\pm$ 0.006 & 3.484 $\pm$ 0.003 & 3.465 $\pm$ 0.020 \\
FineWeb-Edu & 70M  & 8 & \textbf{3.212 $\pm$ 0.016} & 3.282 $\pm$ 0.001 & 3.325 $\pm$ 0.007 & 3.277 $\pm$ 0.010 & 3.323 $\pm$ 0.018 & 3.342 $\pm$ 0.006 \\
FineWeb-Edu & 160M & 4 & \textbf{2.973 $\pm$ 0.008} & 3.028 $\pm$ 0.002 & 3.041 $\pm$ 0.007 & 2.990 $\pm$ 0.003 & 3.061 $\pm$ 0.012 & 3.097 $\pm$ 0.019 \\
FineWeb-Edu & 160M & 8 & \textbf{2.870 $\pm$ 0.006} & 2.930 $\pm$ 0.004 & 2.935 $\pm$ 0.002 & 2.930 $\pm$ 0.003 & 2.935 $\pm$ 0.011 & 2.992 $\pm$ 0.016 \\
\midrule
Lean & 70M  & 4 & \textbf{1.649 $\pm$ 0.012} & 2.115 $\pm$ 0.029 & 2.441 $\pm$ 0.035 & 2.095 $\pm$ 0.034 & 2.202 $\pm$ 0.033 & 2.225 $\pm$ 0.047 \\
Lean & 70M  & 8 & \textbf{1.522 $\pm$ 0.017} & 2.103 $\pm$ 0.022 & 2.323 $\pm$ 0.014 & 1.784 $\pm$ 0.067 & 2.195 $\pm$ 0.028 & 2.175 $\pm$ 0.026 \\
Lean & 160M & 4 & \textbf{1.698 $\pm$ 0.030} & 2.108 $\pm$ 0.013 & 2.545 $\pm$ 0.024 & 1.792 $\pm$ 0.084 & 2.257 $\pm$ 0.013 & 2.229 $\pm$ 0.071 \\
Lean & 160M & 8 & \textbf{1.570 $\pm$ 0.016} & 2.084 $\pm$ 0.016 & 2.407 $\pm$ 0.019 & 1.687 $\pm$ 0.037 & 2.215 $\pm$ 0.026 & 2.185 $\pm$ 0.077 \\
\midrule
DeepMind Mathematics & 70M  & 4 & \textbf{0.473 $\pm$ 0.008} & 0.487 $\pm$ 0.006 & 0.537 $\pm$ 0.003 & 0.556 $\pm$ 0.005 & 0.583 $\pm$ 0.010 & 0.557 $\pm$ 0.005 \\
DeepMind Mathematics & 70M  & 8 & \textbf{0.386 $\pm$ 0.005} & 0.395 $\pm$ 0.003 & 0.511 $\pm$ 0.002 & 0.531 $\pm$ 0.006 & 0.468 $\pm$ 0.004 & 0.462 $\pm$ 0.017 \\
DeepMind Mathematics & 160M & 4 & \textbf{0.383 $\pm$ 0.004} & 0.391 $\pm$ 0.003 & 0.472 $\pm$ 0.010 & 0.740 $\pm$ 0.034 & 0.490 $\pm$ 0.018 & 0.437 $\pm$ 0.015 \\
DeepMind Mathematics & 160M & 8 & 0.270 $\pm$ 0.004 & \textbf{0.269 $\pm$ 0.004} & 0.473 $\pm$ 0.008 & 0.809 $\pm$ 0.035 & 0.347 $\pm$ 0.006 & 0.361 $\pm$ 0.015 \\
\midrule
Stack & 70M  & 4 & \textbf{0.899 $\pm$ 0.003} & 1.376 $\pm$ 0.011 & 1.162 $\pm$ 0.018 & 0.948 $\pm$ 0.004 & 1.256 $\pm$ 0.068 & 1.396 $\pm$ 0.051 \\
Stack & 70M  & 8 & \textbf{0.832 $\pm$ 0.003} & 1.326 $\pm$ 0.018 & 1.068 $\pm$ 0.014 & 0.876 $\pm$ 0.007 & 1.079 $\pm$ 0.015 & 1.179 $\pm$ 0.015 \\
Stack & 160M & 4 & \textbf{0.747 $\pm$ 0.005} & 1.137 $\pm$ 0.001 & 1.040 $\pm$ 0.008 & 0.774 $\pm$ 0.004 & 0.913 $\pm$ 0.012 & 1.060 $\pm$ 0.050 \\
Stack & 160M & 8 & \textbf{0.680 $\pm$ 0.002} & 1.036 $\pm$ 0.009 & 1.250 $\pm$ 0.324 & 0.721 $\pm$ 0.002 & 0.813 $\pm$ 0.012 & 0.927 $\pm$ 0.046 \\
\bottomrule
\end{tabular}
  }
\end{table*}

On development data, full-prefix TANGO has lower mean NLL than every non-TANGO
baseline in all 16 settings and the lowest mean of all six models in 15.  The
exception is mathematics at 160M and depth 8, where Windowed TANGO is lower by
$0.000682\pm0.000723$ NLL, which is a near tie.  Because checkpoints were
selected on development data, the test results in \cref{tab:full-selected-test}
provide the independent comparison.  The higher NLL of Untied Transformer++ on
mathematics at 160M and the high variance of Loop Transformer++ on The Stack at
160M and depth 8 appear in both development and test evaluations.

\section{Official DeepMind Mathematics test evaluation}
\label{app:dm-diagnostics}

This appendix reports the paper's measurements of answer accuracy.  We evaluate
the selected checkpoints on all 560,000 official interpolation examples and all
150,000 official extrapolation examples.  Teacher-forced NLL weights every
answer and end-of-sequence target equally across the full split.  Greedy
generation is scored by exact string equality without normalization, with a
cap of 30 answer characters plus the end-of-sequence symbol.  Every result
includes all three seeds, and averaged over those seeds, at most $0.32\%$ of
interpolation answers and $1.25\%$ of extrapolation answers reach the length
cap, and the table reports this rate for every configuration.

\par\medskip
\noindent\textbf{Answer-plus-EOS test NLL}\par\nobreak\smallskip\nobreak
\noindent\resizebox{\linewidth}{!}{%
\begin{tabular}{lllrrrrrr}
\toprule
Tier & Depth & Split & TANGO & \shortstack{Windowed\\TANGO} & \shortstack{Loop\\Transformer++} & \shortstack{Untied\\Transformer++} & \shortstack{Full-attention\\GAU} & FLASH \\
\midrule
70M & 4 & Interpolation & \textbf{0.554 $\pm$ 0.004} & 0.567 $\pm$ 0.002 & 0.619 $\pm$ 0.004 & 0.637 $\pm$ 0.002 & 0.671 $\pm$ 0.006 & 0.643 $\pm$ 0.009 \\
70M & 4 & Extrapolation & \textbf{0.752 $\pm$ 0.027} & 0.761 $\pm$ 0.011 & 0.857 $\pm$ 0.024 & 0.869 $\pm$ 0.024 & 0.956 $\pm$ 0.022 & 0.901 $\pm$ 0.009 \\
70M & 8 & Interpolation & \textbf{0.463 $\pm$ 0.009} & 0.472 $\pm$ 0.004 & 0.594 $\pm$ 0.001 & 0.607 $\pm$ 0.008 & 0.551 $\pm$ 0.006 & 0.544 $\pm$ 0.014 \\
70M & 8 & Extrapolation & \textbf{0.665 $\pm$ 0.036} & 0.685 $\pm$ 0.018 & 0.822 $\pm$ 0.004 & 0.834 $\pm$ 0.015 & 0.814 $\pm$ 0.031 & 0.813 $\pm$ 0.012 \\
160M & 4 & Interpolation & \textbf{0.460 $\pm$ 0.002} & 0.465 $\pm$ 0.003 & 0.552 $\pm$ 0.011 & 0.830 $\pm$ 0.031 & 0.580 $\pm$ 0.020 & 0.522 $\pm$ 0.018 \\
160M & 4 & Extrapolation & \textbf{0.693 $\pm$ 0.019} & 0.697 $\pm$ 0.024 & 0.768 $\pm$ 0.030 & 1.079 $\pm$ 0.046 & 0.894 $\pm$ 0.048 & 0.799 $\pm$ 0.037 \\
160M & 8 & Interpolation & \textbf{0.340 $\pm$ 0.003} & 0.341 $\pm$ 0.004 & 0.549 $\pm$ 0.007 & 0.896 $\pm$ 0.039 & 0.427 $\pm$ 0.006 & 0.444 $\pm$ 0.014 \\
160M & 8 & Extrapolation & \textbf{0.569 $\pm$ 0.015} & 0.599 $\pm$ 0.022 & 0.735 $\pm$ 0.022 & 1.166 $\pm$ 0.058 & 0.740 $\pm$ 0.020 & 0.768 $\pm$ 0.058 \\
\bottomrule
\end{tabular}
}\par

\par\medskip
\noindent\textbf{Greedy exact answer match (\%)}\par\nobreak\smallskip\nobreak
\noindent\resizebox{\linewidth}{!}{%
\begin{tabular}{lllrrrrrr}
\toprule
Tier & Depth & Split & TANGO & \shortstack{Windowed\\TANGO} & \shortstack{Loop\\Transformer++} & \shortstack{Untied\\Transformer++} & \shortstack{Full-attention\\GAU} & FLASH \\
\midrule
70M & 4 & Interpolation & \textbf{33.42 $\pm$ 0.41} & 32.94 $\pm$ 0.24 & 28.53 $\pm$ 0.36 & 27.57 $\pm$ 0.24 & 26.07 $\pm$ 0.44 & 27.50 $\pm$ 0.41 \\
70M & 4 & Extrapolation & \textbf{28.40 $\pm$ 1.79} & 27.24 $\pm$ 1.02 & 22.04 $\pm$ 0.72 & 19.64 $\pm$ 0.74 & 17.35 $\pm$ 1.14 & 17.80 $\pm$ 0.86 \\
70M & 8 & Interpolation & \textbf{40.09 $\pm$ 0.68} & 39.90 $\pm$ 0.50 & 30.71 $\pm$ 0.09 & 29.67 $\pm$ 0.57 & 33.50 $\pm$ 0.46 & 33.82 $\pm$ 0.93 \\
70M & 8 & Extrapolation & \textbf{33.13 $\pm$ 1.66} & 31.86 $\pm$ 1.41 & 24.99 $\pm$ 1.30 & 23.94 $\pm$ 0.96 & 23.21 $\pm$ 2.19 & 23.12 $\pm$ 1.31 \\
160M & 4 & Interpolation & \textbf{41.21 $\pm$ 0.17} & 40.88 $\pm$ 0.41 & 33.76 $\pm$ 1.20 & 17.82 $\pm$ 1.11 & 32.38 $\pm$ 1.37 & 35.76 $\pm$ 1.04 \\
160M & 4 & Extrapolation & \textbf{34.20 $\pm$ 0.58} & 32.71 $\pm$ 0.34 & 27.47 $\pm$ 1.74 & 10.35 $\pm$ 1.87 & 20.80 $\pm$ 2.58 & 25.53 $\pm$ 1.56 \\
160M & 8 & Interpolation & 51.86 $\pm$ 0.37 & \textbf{52.07 $\pm$ 0.71} & 33.86 $\pm$ 0.28 & 15.10 $\pm$ 1.43 & 43.02 $\pm$ 0.94 & 41.62 $\pm$ 1.30 \\
160M & 8 & Extrapolation & \textbf{40.96 $\pm$ 0.14} & 39.72 $\pm$ 1.68 & 28.46 $\pm$ 0.50 & 7.47 $\pm$ 1.14 & 30.21 $\pm$ 0.75 & 27.93 $\pm$ 2.69 \\
\bottomrule
\end{tabular}
}\par

\par\medskip
\noindent\textbf{Generation cap reached (\%)}\par\nobreak\smallskip\nobreak
\noindent\resizebox{\linewidth}{!}{%
\begin{tabular}{lllrrrrrr}
\toprule
Tier & Depth & Split & TANGO & \shortstack{Windowed\\TANGO} & \shortstack{Loop\\Transformer++} & \shortstack{Untied\\Transformer++} & \shortstack{Full-attention\\GAU} & FLASH \\
\midrule
70M & 4 & Interpolation & \textbf{0.053 $\pm$ 0.015} & 0.062 $\pm$ 0.021 & 0.089 $\pm$ 0.025 & 0.073 $\pm$ 0.030 & 0.080 $\pm$ 0.019 & 0.065 $\pm$ 0.014 \\
70M & 4 & Extrapolation & 0.270 $\pm$ 0.237 & 0.892 $\pm$ 0.652 & 0.465 $\pm$ 0.290 & 0.445 $\pm$ 0.025 & 0.212 $\pm$ 0.050 & \textbf{0.095 $\pm$ 0.052} \\
70M & 8 & Interpolation & \textbf{0.054 $\pm$ 0.007} & 0.058 $\pm$ 0.017 & 0.066 $\pm$ 0.008 & 0.073 $\pm$ 0.009 & 0.071 $\pm$ 0.015 & 0.064 $\pm$ 0.006 \\
70M & 8 & Extrapolation & 0.287 $\pm$ 0.048 & 0.361 $\pm$ 0.125 & 0.626 $\pm$ 0.316 & 0.478 $\pm$ 0.554 & 0.356 $\pm$ 0.251 & \textbf{0.241 $\pm$ 0.304} \\
160M & 4 & Interpolation & \textbf{0.062 $\pm$ 0.027} & 0.072 $\pm$ 0.032 & 0.078 $\pm$ 0.019 & 0.128 $\pm$ 0.059 & 0.070 $\pm$ 0.006 & 0.065 $\pm$ 0.012 \\
160M & 4 & Extrapolation & 0.624 $\pm$ 0.548 & 0.521 $\pm$ 0.420 & \textbf{0.100 $\pm$ 0.047} & 0.465 $\pm$ 0.194 & 0.145 $\pm$ 0.127 & 0.207 $\pm$ 0.046 \\
160M & 8 & Interpolation & \textbf{0.052 $\pm$ 0.002} & 0.063 $\pm$ 0.004 & 0.105 $\pm$ 0.044 & 0.312 $\pm$ 0.239 & 0.063 $\pm$ 0.026 & 0.064 $\pm$ 0.009 \\
160M & 8 & Extrapolation & 0.229 $\pm$ 0.052 & 0.601 $\pm$ 0.102 & 0.233 $\pm$ 0.284 & 1.247 $\pm$ 1.119 & \textbf{0.081 $\pm$ 0.004} & 0.103 $\pm$ 0.090 \\
\bottomrule
\end{tabular}
}\par

TANGO has the lowest mean NLL in all eight combinations of size, depth, and
split and the highest mean exact match in seven of them.  The exception is
interpolation at 160M and depth 8, where Windowed TANGO answers
$52.07\pm0.71\%$ of problems exactly, compared with $51.86\pm0.37\%$ for TANGO.
The two NLLs in that setting differ by less than their variation across seeds,
and on the extrapolation split TANGO has the higher exact match, at
$40.96\pm0.14\%$ versus $39.72\pm1.68\%$.

\subsection{Accuracy by mathematical category}
\label{app:dm-categories}

The following tables break exact match down by mathematical category.  Each
table aggregates examples within each category and reports the mean $\pm$
standard deviation across seeds.  Because each official module contains the
same number of examples, modules receive equal weight within a category.  The
interpolation and extrapolation splits contain different sets of modules, so
their category scores should not be read as paired measurements on the same
problems.

TANGO does not have the highest aggregate accuracy in every category.  On
the probability category, a non-TANGO architecture has the highest mean
accuracy in all eight combinations of size, depth, and split.  At 160M and
depth 4, for example, FLASH answers $9.63\%$ of interpolation and $3.92\%$ of
extrapolation probability problems correctly.

\par\medskip
\noindent\textbf{70M, depth 4, interpolation}\par\nobreak\smallskip\nobreak
\noindent\resizebox{\linewidth}{!}{%
\begin{tabular}{lrrrrrr}
\toprule
Category & TANGO & \shortstack{Windowed\\TANGO} & \shortstack{Loop\\Transformer++} & \shortstack{Untied\\Transformer++} & \shortstack{Full-attention\\GAU} & FLASH \\
\midrule
Algebra & \textbf{16.13 $\pm$ 0.50} & 15.39 $\pm$ 0.60 & 11.70 $\pm$ 0.59 & 11.66 $\pm$ 0.07 & 10.22 $\pm$ 0.64 & 11.14 $\pm$ 0.23 \\
Arithmetic & \textbf{28.39 $\pm$ 0.93} & 27.95 $\pm$ 0.15 & 25.47 $\pm$ 0.27 & 22.14 $\pm$ 0.16 & 15.09 $\pm$ 0.93 & 18.04 $\pm$ 0.41 \\
Calculus & \textbf{57.37 $\pm$ 0.92} & 50.09 $\pm$ 0.72 & 46.39 $\pm$ 0.00 & 40.38 $\pm$ 1.13 & 28.80 $\pm$ 7.25 & 40.39 $\pm$ 2.63 \\
Comparison & 69.25 $\pm$ 0.79 & \textbf{69.67 $\pm$ 0.81} & 57.44 $\pm$ 1.87 & 57.33 $\pm$ 1.13 & 60.79 $\pm$ 0.67 & 60.28 $\pm$ 2.44 \\
Measurement & 45.10 $\pm$ 7.30 & \textbf{46.37 $\pm$ 4.22} & 31.02 $\pm$ 1.69 & 28.87 $\pm$ 0.97 & 28.76 $\pm$ 3.69 & 31.55 $\pm$ 5.06 \\
Numbers & \textbf{37.57 $\pm$ 0.14} & 37.41 $\pm$ 0.30 & 34.90 $\pm$ 0.37 & 34.87 $\pm$ 0.54 & 33.72 $\pm$ 0.77 & 34.72 $\pm$ 0.37 \\
Polynomials & \textbf{9.55 $\pm$ 0.10} & 8.77 $\pm$ 0.21 & 6.77 $\pm$ 0.31 & 6.62 $\pm$ 0.22 & 6.54 $\pm$ 0.25 & 7.45 $\pm$ 0.18 \\
Probability & 6.61 $\pm$ 0.49 & 6.88 $\pm$ 0.49 & 6.49 $\pm$ 0.50 & 4.37 $\pm$ 0.88 & \textbf{7.56 $\pm$ 1.22} & 6.31 $\pm$ 0.08 \\
\bottomrule
\end{tabular}
}\par

\par\medskip
\noindent\textbf{70M, depth 4, extrapolation}\par\nobreak\smallskip\nobreak
\noindent\resizebox{\linewidth}{!}{%
\begin{tabular}{lrrrrrr}
\toprule
Category & TANGO & \shortstack{Windowed\\TANGO} & \shortstack{Loop\\Transformer++} & \shortstack{Untied\\Transformer++} & \shortstack{Full-attention\\GAU} & FLASH \\
\midrule
Algebra & \textbf{2.89 $\pm$ 0.64} & 2.68 $\pm$ 0.63 & 0.87 $\pm$ 0.06 & 0.51 $\pm$ 0.05 & 0.18 $\pm$ 0.09 & 0.57 $\pm$ 0.13 \\
Arithmetic & 15.45 $\pm$ 0.96 & \textbf{15.65 $\pm$ 0.12} & 13.48 $\pm$ 0.23 & 11.57 $\pm$ 0.23 & 6.07 $\pm$ 0.75 & 8.59 $\pm$ 0.51 \\
Comparison & \textbf{40.59 $\pm$ 1.87} & 36.20 $\pm$ 1.23 & 22.86 $\pm$ 2.70 & 21.36 $\pm$ 2.50 & 26.19 $\pm$ 1.65 & 25.41 $\pm$ 3.63 \\
Measurement & 41.42 $\pm$ 5.46 & \textbf{43.24 $\pm$ 0.75} & 23.85 $\pm$ 6.09 & 25.25 $\pm$ 3.18 & 18.05 $\pm$ 9.48 & 27.99 $\pm$ 3.87 \\
Numbers & \textbf{81.08 $\pm$ 9.81} & 76.88 $\pm$ 6.04 & 75.36 $\pm$ 3.90 & 65.06 $\pm$ 9.05 & 59.86 $\pm$ 6.19 & 51.58 $\pm$ 8.58 \\
Probability & 2.50 $\pm$ 0.08 & 3.20 $\pm$ 0.42 & 2.83 $\pm$ 0.31 & 2.61 $\pm$ 0.33 & 3.67 $\pm$ 0.18 & \textbf{3.76 $\pm$ 0.17} \\
\bottomrule
\end{tabular}
}\par

\par\medskip
\noindent\textbf{70M, depth 8, interpolation}\par\nobreak\smallskip\nobreak
\noindent\resizebox{\linewidth}{!}{%
\begin{tabular}{lrrrrrr}
\toprule
Category & TANGO & \shortstack{Windowed\\TANGO} & \shortstack{Loop\\Transformer++} & \shortstack{Untied\\Transformer++} & \shortstack{Full-attention\\GAU} & FLASH \\
\midrule
Algebra & \textbf{25.45 $\pm$ 1.34} & 23.87 $\pm$ 0.66 & 13.32 $\pm$ 0.63 & 12.85 $\pm$ 1.23 & 16.43 $\pm$ 0.36 & 17.56 $\pm$ 1.58 \\
Arithmetic & 39.68 $\pm$ 2.70 & \textbf{39.87 $\pm$ 2.30} & 27.93 $\pm$ 0.90 & 24.61 $\pm$ 1.88 & 28.72 $\pm$ 1.75 & 28.69 $\pm$ 2.45 \\
Calculus & \textbf{66.43 $\pm$ 0.63} & 61.66 $\pm$ 0.57 & 50.68 $\pm$ 1.37 & 47.23 $\pm$ 2.70 & 54.83 $\pm$ 1.27 & 59.42 $\pm$ 3.62 \\
Comparison & 73.15 $\pm$ 0.43 & \textbf{73.55 $\pm$ 1.03} & 63.43 $\pm$ 1.07 & 63.26 $\pm$ 0.73 & 67.79 $\pm$ 1.34 & 68.47 $\pm$ 1.36 \\
Measurement & 65.40 $\pm$ 3.30 & \textbf{67.81 $\pm$ 8.01} & 35.39 $\pm$ 4.93 & 32.55 $\pm$ 3.68 & 50.29 $\pm$ 10.40 & 44.39 $\pm$ 10.16 \\
Numbers & \textbf{42.10 $\pm$ 0.28} & 42.08 $\pm$ 0.22 & 35.93 $\pm$ 0.11 & 35.45 $\pm$ 0.19 & 37.65 $\pm$ 0.04 & 37.93 $\pm$ 0.63 \\
Polynomials & 13.20 $\pm$ 0.12 & \textbf{13.22 $\pm$ 0.46} & 7.45 $\pm$ 0.10 & 7.43 $\pm$ 0.05 & 9.44 $\pm$ 0.11 & 9.90 $\pm$ 0.77 \\
Probability & 7.07 $\pm$ 0.29 & 8.04 $\pm$ 0.42 & 5.92 $\pm$ 0.17 & 4.71 $\pm$ 0.25 & \textbf{8.99 $\pm$ 1.06} & 7.88 $\pm$ 0.12 \\
\bottomrule
\end{tabular}
}\par

\par\medskip
\noindent\textbf{70M, depth 8, extrapolation}\par\nobreak\smallskip\nobreak
\noindent\resizebox{\linewidth}{!}{%
\begin{tabular}{lrrrrrr}
\toprule
Category & TANGO & \shortstack{Windowed\\TANGO} & \shortstack{Loop\\Transformer++} & \shortstack{Untied\\Transformer++} & \shortstack{Full-attention\\GAU} & FLASH \\
\midrule
Algebra & \textbf{6.43 $\pm$ 0.78} & 6.40 $\pm$ 0.85 & 1.47 $\pm$ 0.02 & 1.09 $\pm$ 0.38 & 2.36 $\pm$ 1.25 & 2.55 $\pm$ 1.06 \\
Arithmetic & 21.54 $\pm$ 1.21 & \textbf{21.93 $\pm$ 2.00} & 15.73 $\pm$ 0.36 & 13.31 $\pm$ 0.65 & 13.49 $\pm$ 2.02 & 14.50 $\pm$ 1.57 \\
Comparison & \textbf{49.56 $\pm$ 1.28} & 44.22 $\pm$ 3.99 & 27.60 $\pm$ 3.75 & 26.99 $\pm$ 4.17 & 33.01 $\pm$ 9.60 & 30.79 $\pm$ 8.53 \\
Measurement & 49.53 $\pm$ 6.67 & \textbf{51.91 $\pm$ 3.74} & 44.43 $\pm$ 6.50 & 37.71 $\pm$ 13.64 & 32.93 $\pm$ 10.60 & 33.99 $\pm$ 3.47 \\
Numbers & \textbf{78.46 $\pm$ 8.50} & 74.39 $\pm$ 2.06 & 73.14 $\pm$ 6.16 & 77.32 $\pm$ 6.24 & 62.67 $\pm$ 7.82 & 61.51 $\pm$ 2.92 \\
Probability & 3.07 $\pm$ 0.33 & 3.33 $\pm$ 0.25 & 2.78 $\pm$ 0.48 & 2.46 $\pm$ 0.19 & 3.74 $\pm$ 0.34 & \textbf{3.94 $\pm$ 0.29} \\
\bottomrule
\end{tabular}
}\par

\par\medskip
\noindent\textbf{160M, depth 4, interpolation}\par\nobreak\smallskip\nobreak
\noindent\resizebox{\linewidth}{!}{%
\begin{tabular}{lrrrrrr}
\toprule
Category & TANGO & \shortstack{Windowed\\TANGO} & \shortstack{Loop\\Transformer++} & \shortstack{Untied\\Transformer++} & \shortstack{Full-attention\\GAU} & FLASH \\
\midrule
Algebra & 27.04 $\pm$ 0.91 & \textbf{27.27 $\pm$ 0.51} & 18.49 $\pm$ 1.63 & 6.56 $\pm$ 0.66 & 15.60 $\pm$ 1.88 & 19.84 $\pm$ 1.28 \\
Arithmetic & \textbf{40.91 $\pm$ 0.95} & 40.78 $\pm$ 1.03 & 31.41 $\pm$ 1.37 & 8.34 $\pm$ 1.82 & 26.00 $\pm$ 1.76 & 30.93 $\pm$ 1.67 \\
Calculus & \textbf{66.22 $\pm$ 0.35} & 64.08 $\pm$ 0.66 & 57.21 $\pm$ 1.38 & 7.40 $\pm$ 1.99 & 52.30 $\pm$ 3.79 & 57.49 $\pm$ 5.41 \\
Comparison & \textbf{74.59 $\pm$ 0.97} & 74.29 $\pm$ 0.60 & 66.89 $\pm$ 3.32 & 41.19 $\pm$ 2.44 & 67.16 $\pm$ 1.72 & 70.75 $\pm$ 1.42 \\
Measurement & \textbf{82.31 $\pm$ 2.00} & 78.25 $\pm$ 6.66 & 42.63 $\pm$ 5.57 & 8.34 $\pm$ 1.78 & 52.62 $\pm$ 5.41 & 60.21 $\pm$ 3.42 \\
Numbers & \textbf{41.74 $\pm$ 0.03} & 41.49 $\pm$ 0.33 & 37.55 $\pm$ 0.22 & 27.92 $\pm$ 1.00 & 36.42 $\pm$ 0.85 & 38.73 $\pm$ 0.98 \\
Polynomials & \textbf{13.17 $\pm$ 0.32} & 12.89 $\pm$ 0.16 & 9.30 $\pm$ 0.21 & 3.80 $\pm$ 0.10 & 9.18 $\pm$ 1.22 & 10.79 $\pm$ 0.88 \\
Probability & 7.39 $\pm$ 0.23 & 8.28 $\pm$ 0.42 & 6.14 $\pm$ 0.46 & 2.10 $\pm$ 0.53 & 7.44 $\pm$ 0.21 & \textbf{9.63 $\pm$ 1.13} \\
\bottomrule
\end{tabular}
}\par

\par\medskip
\noindent\textbf{160M, depth 4, extrapolation}\par\nobreak\smallskip\nobreak
\noindent\resizebox{\linewidth}{!}{%
\begin{tabular}{lrrrrrr}
\toprule
Category & TANGO & \shortstack{Windowed\\TANGO} & \shortstack{Loop\\Transformer++} & \shortstack{Untied\\Transformer++} & \shortstack{Full-attention\\GAU} & FLASH \\
\midrule
Algebra & \textbf{7.05 $\pm$ 1.46} & 5.21 $\pm$ 1.35 & 2.41 $\pm$ 1.02 & 0.21 $\pm$ 0.08 & 1.10 $\pm$ 0.45 & 1.65 $\pm$ 0.16 \\
Arithmetic & \textbf{21.32 $\pm$ 0.29} & 20.82 $\pm$ 0.78 & 16.71 $\pm$ 0.67 & 4.64 $\pm$ 0.90 & 9.86 $\pm$ 1.03 & 13.73 $\pm$ 1.49 \\
Comparison & \textbf{52.08 $\pm$ 3.89} & 46.37 $\pm$ 1.12 & 33.56 $\pm$ 7.89 & 11.01 $\pm$ 0.40 & 31.55 $\pm$ 7.40 & 36.90 $\pm$ 5.43 \\
Measurement & \textbf{57.88 $\pm$ 7.74} & 55.96 $\pm$ 7.38 & 43.32 $\pm$ 3.29 & 4.98 $\pm$ 1.92 & 22.61 $\pm$ 6.45 & 35.14 $\pm$ 6.18 \\
Numbers & 79.15 $\pm$ 4.61 & \textbf{79.44 $\pm$ 2.64} & 79.06 $\pm$ 2.35 & 42.63 $\pm$ 11.06 & 63.36 $\pm$ 5.23 & 72.62 $\pm$ 3.14 \\
Probability & 2.79 $\pm$ 0.28 & 3.28 $\pm$ 0.77 & 3.63 $\pm$ 0.38 & 1.98 $\pm$ 0.12 & 3.86 $\pm$ 0.54 & \textbf{3.92 $\pm$ 0.10} \\
\bottomrule
\end{tabular}
}\par

\par\medskip
\noindent\textbf{160M, depth 8, interpolation}\par\nobreak\smallskip\nobreak
\noindent\resizebox{\linewidth}{!}{%
\begin{tabular}{lrrrrrr}
\toprule
Category & TANGO & \shortstack{Windowed\\TANGO} & \shortstack{Loop\\Transformer++} & \shortstack{Untied\\Transformer++} & \shortstack{Full-attention\\GAU} & FLASH \\
\midrule
Algebra & \textbf{45.64 $\pm$ 0.80} & 44.86 $\pm$ 0.42 & 18.06 $\pm$ 1.51 & 5.69 $\pm$ 0.60 & 30.05 $\pm$ 1.36 & 29.23 $\pm$ 1.29 \\
Arithmetic & 58.72 $\pm$ 1.87 & \textbf{59.12 $\pm$ 1.05} & 31.64 $\pm$ 0.52 & 5.39 $\pm$ 0.72 & 44.10 $\pm$ 2.11 & 40.65 $\pm$ 3.04 \\
Calculus & \textbf{77.26 $\pm$ 0.58} & 74.13 $\pm$ 1.62 & 55.72 $\pm$ 1.96 & 5.56 $\pm$ 2.24 & 69.44 $\pm$ 2.57 & 69.09 $\pm$ 1.43 \\
Comparison & 81.29 $\pm$ 0.79 & \textbf{81.43 $\pm$ 0.21} & 66.77 $\pm$ 0.46 & 37.52 $\pm$ 2.60 & 76.17 $\pm$ 0.53 & 75.57 $\pm$ 0.70 \\
Measurement & 77.41 $\pm$ 1.26 & \textbf{84.92 $\pm$ 7.01} & 45.91 $\pm$ 2.94 & 4.68 $\pm$ 2.86 & 74.34 $\pm$ 4.36 & 72.21 $\pm$ 3.74 \\
Numbers & \textbf{48.81 $\pm$ 0.18} & 48.69 $\pm$ 0.27 & 37.67 $\pm$ 0.41 & 23.68 $\pm$ 2.07 & 42.89 $\pm$ 1.01 & 41.40 $\pm$ 1.33 \\
Polynomials & 25.52 $\pm$ 0.63 & \textbf{26.14 $\pm$ 1.61} & 9.84 $\pm$ 0.17 & 3.21 $\pm$ 0.62 & 15.57 $\pm$ 1.01 & 15.16 $\pm$ 0.67 \\
Probability & 8.58 $\pm$ 0.70 & 9.36 $\pm$ 0.78 & 5.08 $\pm$ 0.61 & 1.34 $\pm$ 0.24 & \textbf{10.52 $\pm$ 1.10} & 9.53 $\pm$ 0.44 \\
\bottomrule
\end{tabular}
}\par

\par\medskip
\noindent\textbf{160M, depth 8, extrapolation}\par\nobreak\smallskip\nobreak
\noindent\resizebox{\linewidth}{!}{%
\begin{tabular}{lrrrrrr}
\toprule
Category & TANGO & \shortstack{Windowed\\TANGO} & \shortstack{Loop\\Transformer++} & \shortstack{Untied\\Transformer++} & \shortstack{Full-attention\\GAU} & FLASH \\
\midrule
Algebra & \textbf{13.03 $\pm$ 0.80} & 11.15 $\pm$ 1.42 & 2.67 $\pm$ 0.69 & 0.10 $\pm$ 0.03 & 4.96 $\pm$ 0.43 & 4.97 $\pm$ 0.70 \\
Arithmetic & 31.28 $\pm$ 0.36 & \textbf{31.28 $\pm$ 0.30} & 17.89 $\pm$ 0.65 & 2.97 $\pm$ 0.74 & 20.09 $\pm$ 1.33 & 18.86 $\pm$ 1.22 \\
Comparison & \textbf{58.00 $\pm$ 1.41} & 51.05 $\pm$ 2.98 & 36.32 $\pm$ 2.85 & 9.72 $\pm$ 0.74 & 40.82 $\pm$ 2.40 & 33.70 $\pm$ 3.83 \\
Measurement & 73.94 $\pm$ 0.92 & \textbf{74.21 $\pm$ 0.54} & 43.59 $\pm$ 0.35 & 3.00 $\pm$ 2.66 & 52.65 $\pm$ 12.29 & 37.07 $\pm$ 10.33 \\
Numbers & 80.06 $\pm$ 2.53 & \textbf{81.64 $\pm$ 6.55} & 79.26 $\pm$ 1.26 & 29.30 $\pm$ 3.86 & 72.74 $\pm$ 11.14 & 77.58 $\pm$ 8.67 \\
Probability & 2.81 $\pm$ 0.38 & 3.15 $\pm$ 0.35 & 2.93 $\pm$ 0.32 & 1.71 $\pm$ 0.18 & 3.54 $\pm$ 0.54 & \textbf{3.73 $\pm$ 0.06} \\
\bottomrule
\end{tabular}
}\par

\subsection{Relation to published mathematics results}

Our absolute accuracies are well below published results on this benchmark.
Training exposure is one substantial difference between our setting and
theirs.  The original 30M-parameter
Transformer reports $76.00\%$ official interpolation accuracy after 500,000
updates with batch 1,024, which is 512 million example
presentations~\citep{saxton2019analytic}.  The 49.2M-parameter Tensor-Product
Transformer reports $84.24\%$ after 1.7 million updates at the same batch size,
approximately 1.74 billion presentations~\citep{schlag2019tp}.  The
abductive-learning system ABL-Sym uses a Tensor-Product Transformer generator
together with search and evaluation in the SymPy symbolic algebra system, and
it reports $88.52\%$ after 700,000 updates, or 716.8 million
presentations~\citep{hu2022ablsym}.  Our 160M models, by contrast, see 12.50
million examples.  Differences in training exposure, model size, decoding, and
external symbolic computation mean that these numbers do not provide a matched
comparison of sample efficiency or support a state-of-the-art claim.

\section{Compact-controller efficiency experiment}
\label{app:compact-controller}

TANGO with three gated-projection sets reduces the cost of full-prefix TANGO through narrower
active gated projections.  This appendix reports a different approach, which narrows only the
quantity that is averaged over pairs of tokens.  Full-prefix TANGO averages a
gate of width $F$ over every visible pair of source and destination, which is
expensive at long contexts.  The compact-controller variant instead constructs
a smaller nonlinear controller at each source, averages the controller, and
only then expands it into a full gate,
\begin{equation}
 c_{ja}=\operatorname{SiLU}(C_a x_j),\qquad
 \bar c_{ia}=\sum_{j\in\mathcal C_i}\alpha_{ija}c_{ja},\qquad
 \bar g_{ia}=B_a\bar c_{ia}.
 \label{eq:compact-controller}
\end{equation}
Here $c_{ja}$ has 32 components per head, for 512 averaged components in the
70M model, and $\alpha_{ija}$ is the same normalized TANGO attention weight,
including the null option.  The expanded controller $\bar g_i$ then gates the
destination features exactly as in TANGO.  The parameters saved
by averaging the smaller controller are reallocated to the gated projection,
so the non-embedding parameter target remains matched.  This variant is a
different architecture rather than a faster implementation of TANGO, because
it constrains the averaged gates to a learned low-dimensional subspace.

\begin{table*}[ht]
  \centering
  \caption{Compact-controller TANGO at 70M and depth 4.  Quality is
  development NLL at the checkpoint selected from the evaluation schedule,
  reported as the mean $\pm$ standard deviation over three paired seeds, and
  the paired $\Delta$ is compact-controller TANGO minus TANGO, so positive
  values favor TANGO.  Systems measurements use matched complete optimizer updates
  with eager bfloat16 execution on prepared batches, on one H100 graphics
  processor with 80 gigabytes of memory.  Speedup is compact-controller
  throughput divided by TANGO throughput, memory reduction refers to
  peak allocated graphics-processor memory, and ``not measured'' marks a task
  shape that was not measured.}
  \label{tab:compact-controller}
  \normalsize
  \resizebox{\textwidth}{!}{%
\begin{tabular}{lrrrr}
\toprule
Dataset & TANGO & Low-dimensional variant & Paired $\Delta$ & \shortstack{Throughput speedup /\\peak GPU-memory reduction} \\
\midrule
FineWeb-Edu & $3.2931 \pm 0.0046$ & $3.3668 \pm 0.0044$ & $0.0736 \pm 0.0089$ & $2.71\times$ / $8.2\%$ \\
Stack & $0.8995 \pm 0.0032$ & $0.9619 \pm 0.0055$ & $0.0624 \pm 0.0056$ & not measured \\
Lean & $1.6489 \pm 0.0116$ & $1.8445 \pm 0.0035$ & $0.1955 \pm 0.0097$ & not measured \\
DeepMind Mathematics & $0.4733 \pm 0.0081$ & $0.5014 \pm 0.0029$ & $0.0281 \pm 0.0109$ & $1.92\times$ / $7.0\%$ \\
\bottomrule
\end{tabular}
  }
\end{table*}

The compact controller has higher throughput in both measured shapes, but it
has higher development NLL than TANGO on all four datasets.  At
70M and depth 4, it has lower NLL than every non-TANGO baseline in the primary
comparison on Lean and DeepMind Mathematics, but not on FineWeb-Edu or The
Stack.  It is also still slower than Untied Transformer++, with prepared-batch
throughput of approximately 50\% of the untied model's on the FineWeb-Edu shape
and 24\% on the mathematics shape.

\section{Stack test results by programming language}
\label{app:stack-languages}

This appendix breaks the test results on The Stack down by programming
language, using the selected checkpoints for all three seeds and all four
combinations of size and depth.  The test set contains 3,072 files from 1,837
held-out repository families, with 512 files per language.  Because the GPT-2
tokenizer encodes some languages more efficiently than others, we report bits
per byte of the original UTF-8 source (BPB) rather than NLL per token.  For each
file, token NLL in nats is summed and divided by the number of original UTF-8
source bytes times $\log 2$.  The initial context-only token and right padding
contribute neither targets nor bytes.  Each entry is a mean and
standard deviation over three seeds.

\par\medskip
\noindent\textbf{70M, depth 4}\par\nobreak\smallskip\nobreak
\noindent\resizebox{\linewidth}{!}{%
\begin{tabular}{lrrrrrr}
\toprule
Language & TANGO & \shortstack{Windowed\\TANGO} & \shortstack{Loop\\Transformer++} & \shortstack{Untied\\Transformer++} & \shortstack{Full-attention\\GAU} & FLASH \\
\midrule
Python & \textbf{0.654 $\pm$ 0.003} & 0.941 $\pm$ 0.008 & 0.832 $\pm$ 0.012 & 0.691 $\pm$ 0.003 & 0.882 $\pm$ 0.040 & 0.962 $\pm$ 0.030 \\
Java & \textbf{0.461 $\pm$ 0.002} & 0.698 $\pm$ 0.006 & 0.598 $\pm$ 0.009 & 0.488 $\pm$ 0.002 & 0.641 $\pm$ 0.033 & 0.711 $\pm$ 0.028 \\
JavaScript & \textbf{0.633 $\pm$ 0.003} & 0.933 $\pm$ 0.006 & 0.813 $\pm$ 0.012 & 0.668 $\pm$ 0.002 & 0.869 $\pm$ 0.043 & 0.952 $\pm$ 0.031 \\
C++ & \textbf{0.614 $\pm$ 0.002} & 0.957 $\pm$ 0.008 & 0.797 $\pm$ 0.013 & 0.649 $\pm$ 0.003 & 0.866 $\pm$ 0.050 & 0.966 $\pm$ 0.039 \\
Go & \textbf{0.584 $\pm$ 0.003} & 0.956 $\pm$ 0.006 & 0.769 $\pm$ 0.012 & 0.617 $\pm$ 0.003 & 0.857 $\pm$ 0.054 & 0.971 $\pm$ 0.039 \\
Rust & \textbf{0.573 $\pm$ 0.002} & 0.864 $\pm$ 0.007 & 0.735 $\pm$ 0.010 & 0.604 $\pm$ 0.003 & 0.784 $\pm$ 0.039 & 0.867 $\pm$ 0.030 \\
\bottomrule
\end{tabular}
}\par

\par\medskip
\noindent\textbf{70M, depth 8}\par\nobreak\smallskip\nobreak
\noindent\resizebox{\linewidth}{!}{%
\begin{tabular}{lrrrrrr}
\toprule
Language & TANGO & \shortstack{Windowed\\TANGO} & \shortstack{Loop\\Transformer++} & \shortstack{Untied\\Transformer++} & \shortstack{Full-attention\\GAU} & FLASH \\
\midrule
Python & \textbf{0.609 $\pm$ 0.002} & 0.908 $\pm$ 0.011 & 0.772 $\pm$ 0.008 & 0.640 $\pm$ 0.005 & 0.773 $\pm$ 0.009 & 0.831 $\pm$ 0.009 \\
Java & \textbf{0.428 $\pm$ 0.002} & 0.672 $\pm$ 0.010 & 0.553 $\pm$ 0.006 & 0.451 $\pm$ 0.004 & 0.554 $\pm$ 0.007 & 0.599 $\pm$ 0.008 \\
JavaScript & \textbf{0.588 $\pm$ 0.002} & 0.900 $\pm$ 0.010 & 0.749 $\pm$ 0.010 & 0.619 $\pm$ 0.004 & 0.755 $\pm$ 0.011 & 0.816 $\pm$ 0.009 \\
C++ & \textbf{0.567 $\pm$ 0.002} & 0.922 $\pm$ 0.013 & 0.732 $\pm$ 0.011 & 0.597 $\pm$ 0.004 & 0.741 $\pm$ 0.011 & 0.809 $\pm$ 0.011 \\
Go & \textbf{0.536 $\pm$ 0.002} & 0.926 $\pm$ 0.012 & 0.700 $\pm$ 0.010 & 0.566 $\pm$ 0.004 & 0.717 $\pm$ 0.011 & 0.805 $\pm$ 0.013 \\
Rust & \textbf{0.531 $\pm$ 0.002} & 0.829 $\pm$ 0.014 & 0.678 $\pm$ 0.008 & 0.559 $\pm$ 0.005 & 0.681 $\pm$ 0.009 & 0.738 $\pm$ 0.010 \\
\bottomrule
\end{tabular}
}\par

\par\medskip
\noindent\textbf{160M, depth 4}\par\nobreak\smallskip\nobreak
\noindent\resizebox{\linewidth}{!}{%
\begin{tabular}{lrrrrrr}
\toprule
Language & TANGO & \shortstack{Windowed\\TANGO} & \shortstack{Loop\\Transformer++} & \shortstack{Untied\\Transformer++} & \shortstack{Full-attention\\GAU} & FLASH \\
\midrule
Python & \textbf{0.548 $\pm$ 0.004} & 0.791 $\pm$ 0.001 & 0.749 $\pm$ 0.005 & 0.569 $\pm$ 0.003 & 0.660 $\pm$ 0.007 & 0.750 $\pm$ 0.030 \\
Java & \textbf{0.380 $\pm$ 0.002} & 0.570 $\pm$ 0.001 & 0.530 $\pm$ 0.004 & 0.394 $\pm$ 0.002 & 0.462 $\pm$ 0.006 & 0.531 $\pm$ 0.025 \\
JavaScript & \textbf{0.528 $\pm$ 0.004} & 0.775 $\pm$ 0.000 & 0.725 $\pm$ 0.006 & 0.547 $\pm$ 0.003 & 0.640 $\pm$ 0.008 & 0.732 $\pm$ 0.030 \\
C++ & \textbf{0.513 $\pm$ 0.004} & 0.790 $\pm$ 0.001 & 0.721 $\pm$ 0.006 & 0.532 $\pm$ 0.003 & 0.633 $\pm$ 0.008 & 0.735 $\pm$ 0.037 \\
Go & \textbf{0.482 $\pm$ 0.004} & 0.790 $\pm$ 0.002 & 0.689 $\pm$ 0.005 & 0.499 $\pm$ 0.003 & 0.600 $\pm$ 0.009 & 0.721 $\pm$ 0.041 \\
Rust & \textbf{0.473 $\pm$ 0.003} & 0.707 $\pm$ 0.001 & 0.652 $\pm$ 0.005 & 0.491 $\pm$ 0.002 & 0.572 $\pm$ 0.008 & 0.657 $\pm$ 0.029 \\
\bottomrule
\end{tabular}
}\par

\par\medskip
\noindent\textbf{160M, depth 8}\par\nobreak\smallskip\nobreak
\noindent\resizebox{\linewidth}{!}{%
\begin{tabular}{lrrrrrr}
\toprule
Language & TANGO & \shortstack{Windowed\\TANGO} & \shortstack{Loop\\Transformer++} & \shortstack{Untied\\Transformer++} & \shortstack{Full-attention\\GAU} & FLASH \\
\midrule
Python & \textbf{0.503 $\pm$ 0.002} & 0.728 $\pm$ 0.005 & 0.874 $\pm$ 0.191 & 0.532 $\pm$ 0.001 & 0.594 $\pm$ 0.008 & 0.667 $\pm$ 0.029 \\
Java & \textbf{0.347 $\pm$ 0.001} & 0.518 $\pm$ 0.004 & 0.636 $\pm$ 0.164 & 0.368 $\pm$ 0.001 & 0.414 $\pm$ 0.006 & 0.466 $\pm$ 0.023 \\
JavaScript & \textbf{0.483 $\pm$ 0.001} & 0.710 $\pm$ 0.006 & 0.851 $\pm$ 0.192 & 0.511 $\pm$ 0.001 & 0.572 $\pm$ 0.008 & 0.646 $\pm$ 0.029 \\
C++ & \textbf{0.466 $\pm$ 0.001} & 0.717 $\pm$ 0.007 & 0.883 $\pm$ 0.250 & 0.495 $\pm$ 0.001 & 0.562 $\pm$ 0.008 & 0.642 $\pm$ 0.033 \\
Go & \textbf{0.436 $\pm$ 0.001} & 0.720 $\pm$ 0.006 & 0.852 $\pm$ 0.254 & 0.462 $\pm$ 0.001 & 0.529 $\pm$ 0.009 & 0.624 $\pm$ 0.038 \\
Rust & \textbf{0.431 $\pm$ 0.001} & 0.642 $\pm$ 0.005 & 0.767 $\pm$ 0.175 & 0.457 $\pm$ 0.001 & 0.512 $\pm$ 0.007 & 0.578 $\pm$ 0.027 \\
\bottomrule
\end{tabular}
}\par

TANGO has the lowest mean BPB in all 24 combinations of language, size, and
depth.  The difference from Untied Transformer++ occurs in every language in
this collection.  Windowed TANGO has lower mean BPB than FLASH in every
language at 70M and depth 4, but FLASH has lower mean BPB in every language in
the remaining three settings.  The aggregate token-NLL ordering on The Stack
therefore occurs in every language rather than arising from one language.  As
in the main text, these
likelihood comparisons do not measure whether generated programs run
correctly.

\end{document}